\documentclass[10pt,twocolumn,letterpaper]{article}

\usepackage{cvpr}
\usepackage{times}
\usepackage{epsfig}
\usepackage{graphicx}
\usepackage{amsmath}
\usepackage{amssymb}
\usepackage{amsthm}
\usepackage{multirow}
\usepackage{enumerate}

\newtheorem{lemma}{lemma}
\newcommand{\FT}{\mathcal{F}}
\newcommand{\calP}{\mathcal{P}}
\newcommand{\calF}{\mathcal{F}}
\newcommand{\calA}{\mathcal{A}}
\newcommand{\vB}{\mathbf{B}}
\newcommand{\vk}{\mathbf{k}}
\newcommand{\vL}{\mathbf{L}}
\newcommand{\vl}{\mathbf{l}}
\newcommand{\vI}{\mathbf{I}}
\newcommand{\vJ}{\mathbf{J}}
\newcommand{\Tk}{\mathbf{T}_{\vk}}
\newcommand{\fig}[1]{Fig.~\ref{fig:#1}}

\newcommand{\sinc}{{\rm sinc}}
\newcommand{\vc}{{\rm vec}}
\newcommand{\argmin}{{\rm argmin}}

\newcommand{\SKIP}[1]{}

\usepackage[pagebackref=true,breaklinks=true,letterpaper=true,colorlinks,bookmarks=false]{hyperref}

\cvprfinalcopy 

\def\cvprPaperID{2485} 

\ifcvprfinal\fi
 
\begin{document}

\title{Phase-only Image Based Kernel Estimation for Single Image Blind
Deblurring}

\author{Liyuan Pan$^{1,2}$, Richard Hartley$^{1,2}$, Miaomiao Liu$^{1,2}$, and Yuchao Dai$^{3}$\\ 
$^{1}$ Australian National University, Canberra, Australia \\ 
$^{2}$ Australian Centre for Robotic Vision \\
$^{3}$ School of Electronics and Information, Northwestern Polytechnical University, Xi'an, China \\
\tt\small{\{liyuan.pan, Richard.Hartley, miaomiao.liu\}}@anu.edu.au, 
daiyuchao@nwpu.edu.cn
}

\maketitle
\thispagestyle{empty}

\begin{abstract}
The image motion blurring process is generally modelled as the convolution of a blur kernel with a latent image. Therefore, the estimation of the blur kernel is essentially important for blind image deblurring. Unlike existing approaches which focus on approaching the problem by enforcing various priors on the blur kernel and the latent image, we are aiming at obtaining a high quality blur kernel directly by studying the problem in the frequency domain. We show that the auto-correlation of the absolute~\emph{phase-only image}%
\footnote{Phase-only image means the image is reconstructed only from the phase information of the blurry image.} 
can provide faithful information about the motion (\eg, the motion direction and magnitude, we call it the ~\emph{motion pattern} in this paper.) that caused the blur, leading to a new and efficient blur kernel estimation approach. The blur kernel is then refined and the sharp image is estimated by solving an optimization problem by enforcing a regularization on the blur kernel and the latent image. We further extend our approach to handle non-uniform blur, which involves spatially varying blur kernels. Our approach is evaluated extensively on synthetic and real data and shows good results compared to the state-of-the-art deblurring approaches.　
\end{abstract}


\section{Introduction}

\begin{figure*}[t]
\begin{center}
\begin{tabular}[t]{cccc}
\hspace{-0.25 cm}
\multirow{3}{*}{\includegraphics[width=0.213\textwidth]{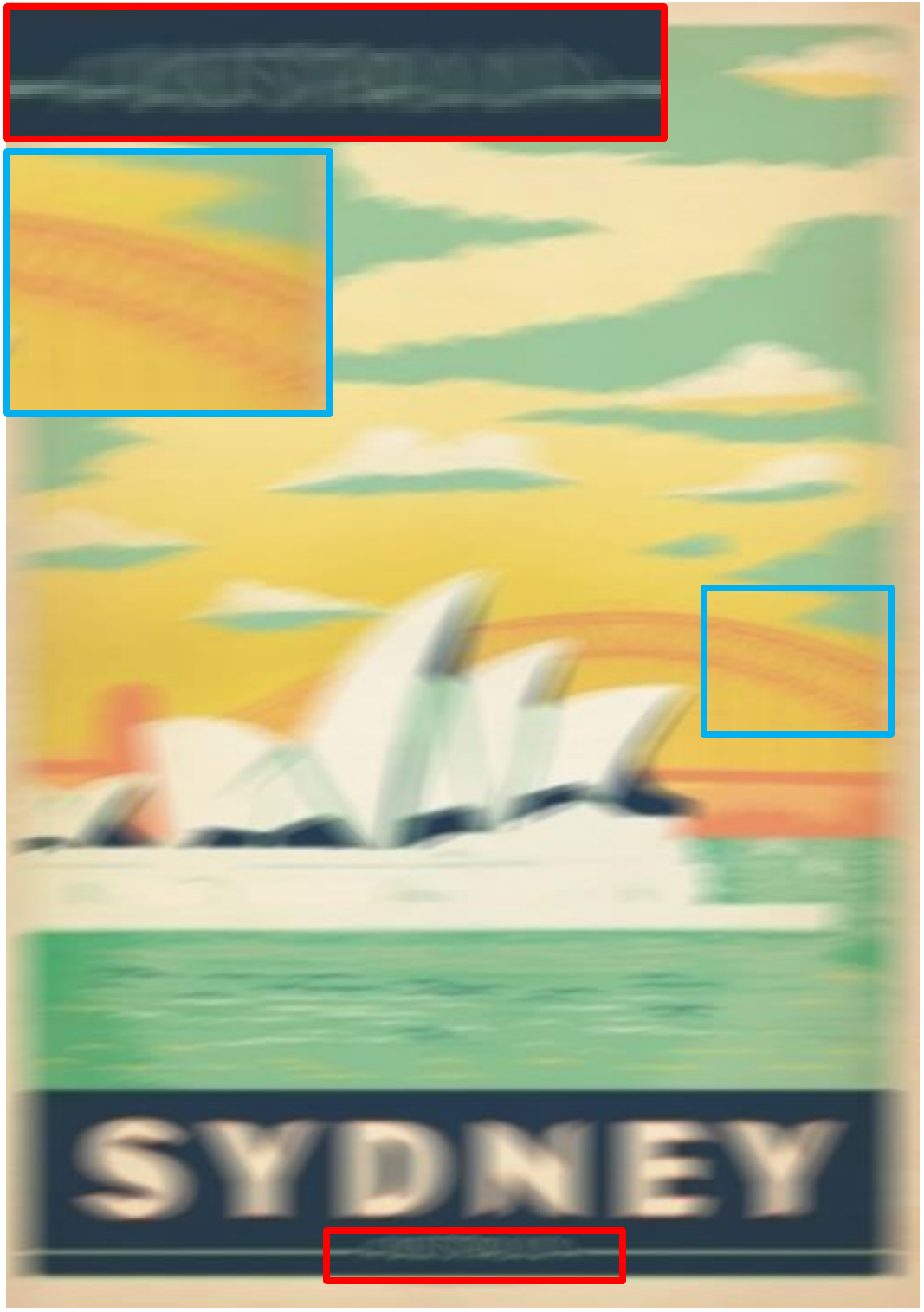}} 
& \multirow{3}{*}{\includegraphics[width=0.213\textwidth]{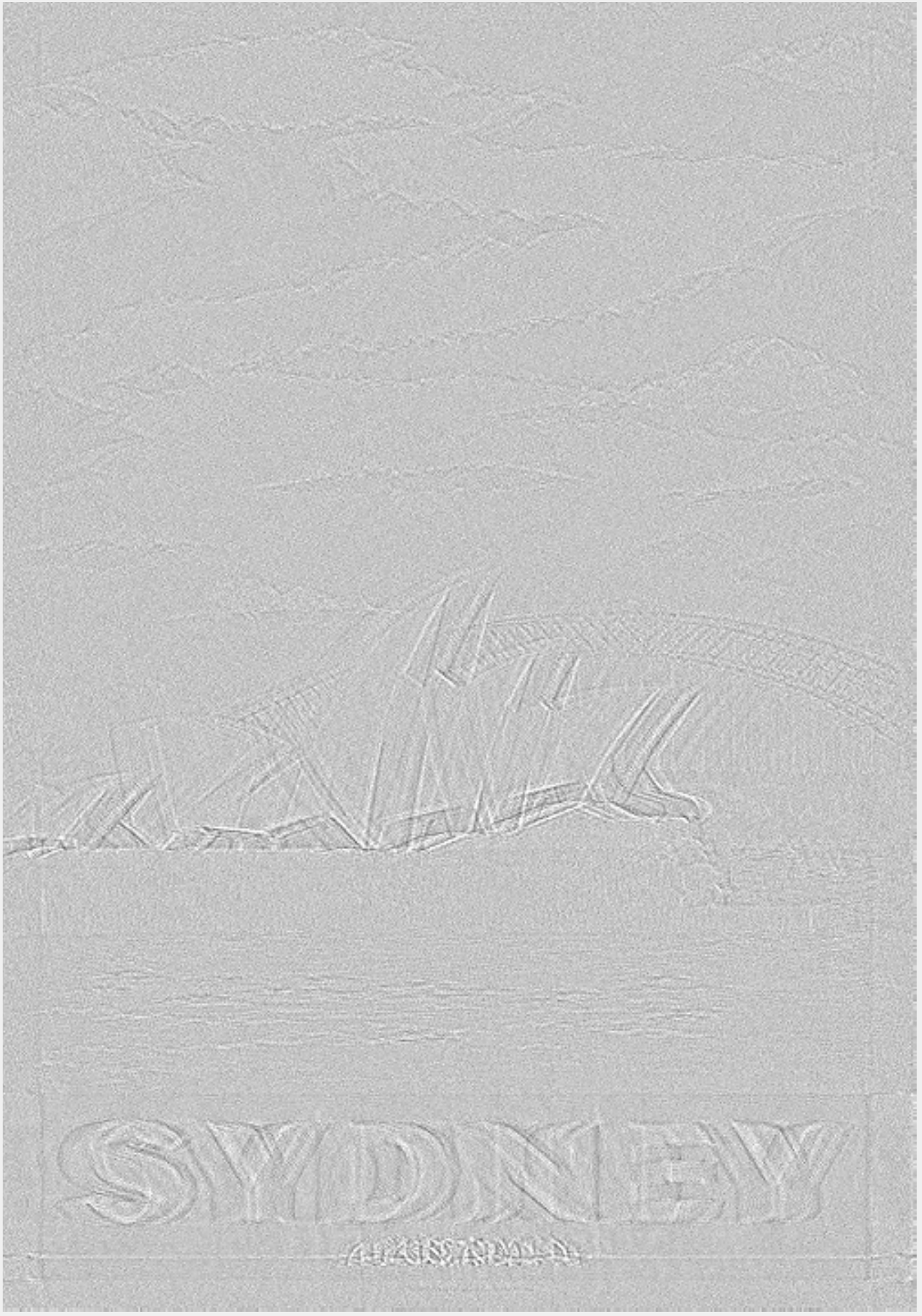}} 
& \raisebox{-2.0cm}{\includegraphics[width=0.213\textwidth]{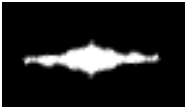}}
& \multirow{3}{*}{\includegraphics[width=0.213\textwidth]{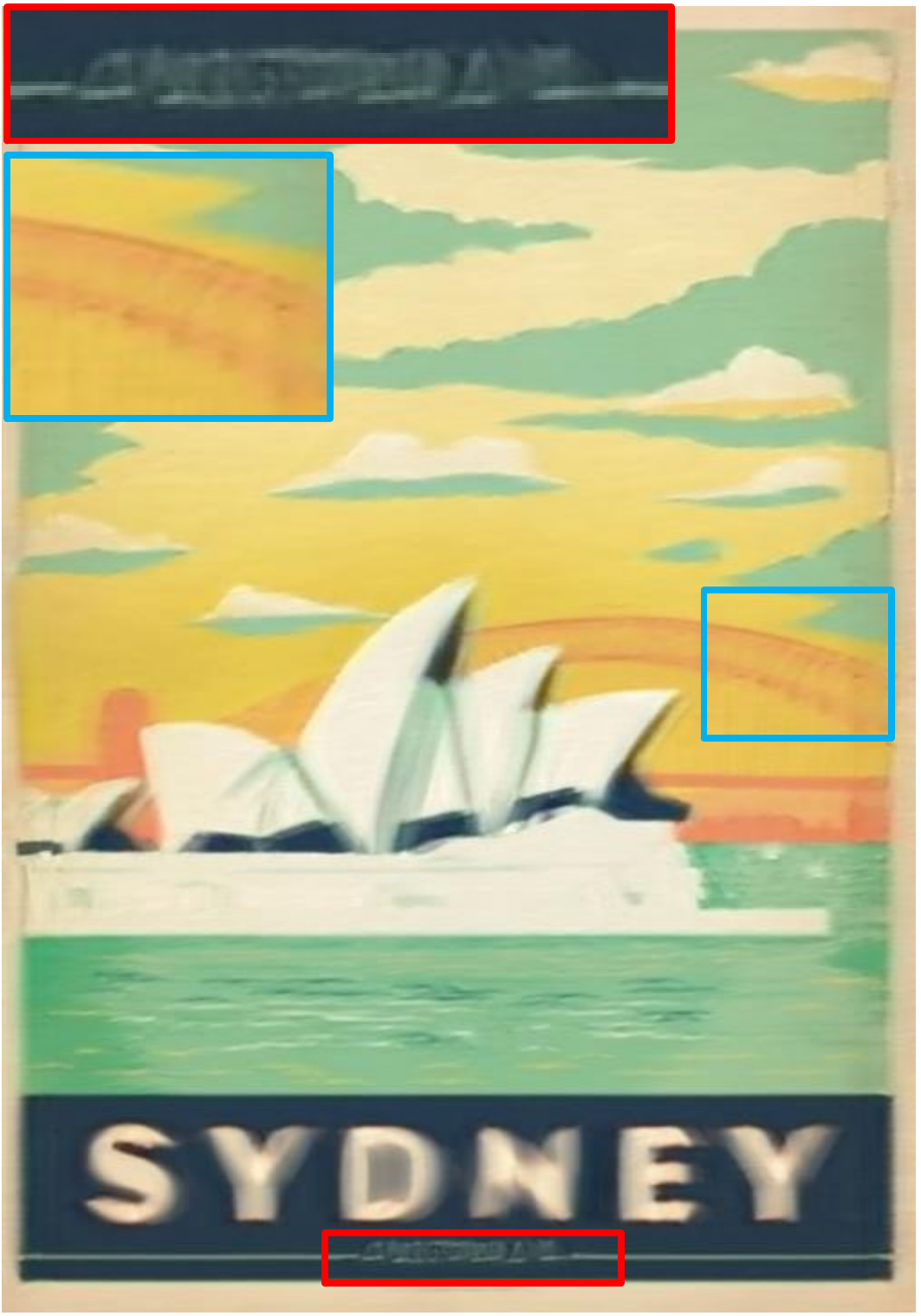}} \\
                   &                    & \raisebox{-0.25 cm}{(c) Auto-correlation}\\
                   &                    &
\raisebox{0 cm}{\includegraphics[width=0.215\textwidth]{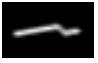}}\\
\hspace{-0.25 cm}
 (a) Blurry Image  
& (b) $P(B)$  
& (d) Kernel 
& (e) Nah~\cite{Nah_2017_CVPR} \\
\hspace{-0.25 cm}
\includegraphics[width=0.213\textwidth]{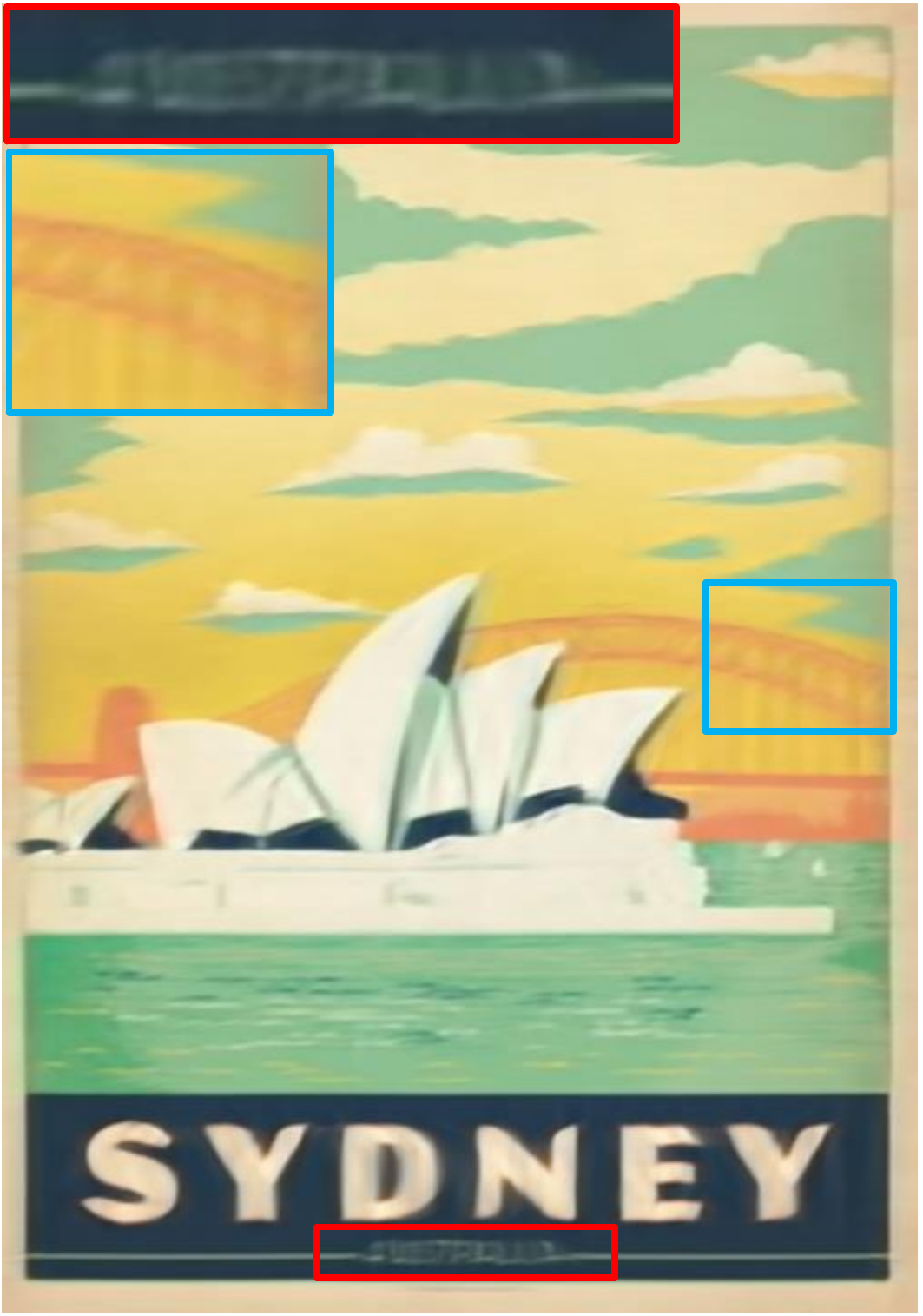}
&\includegraphics[width=0.213\textwidth]{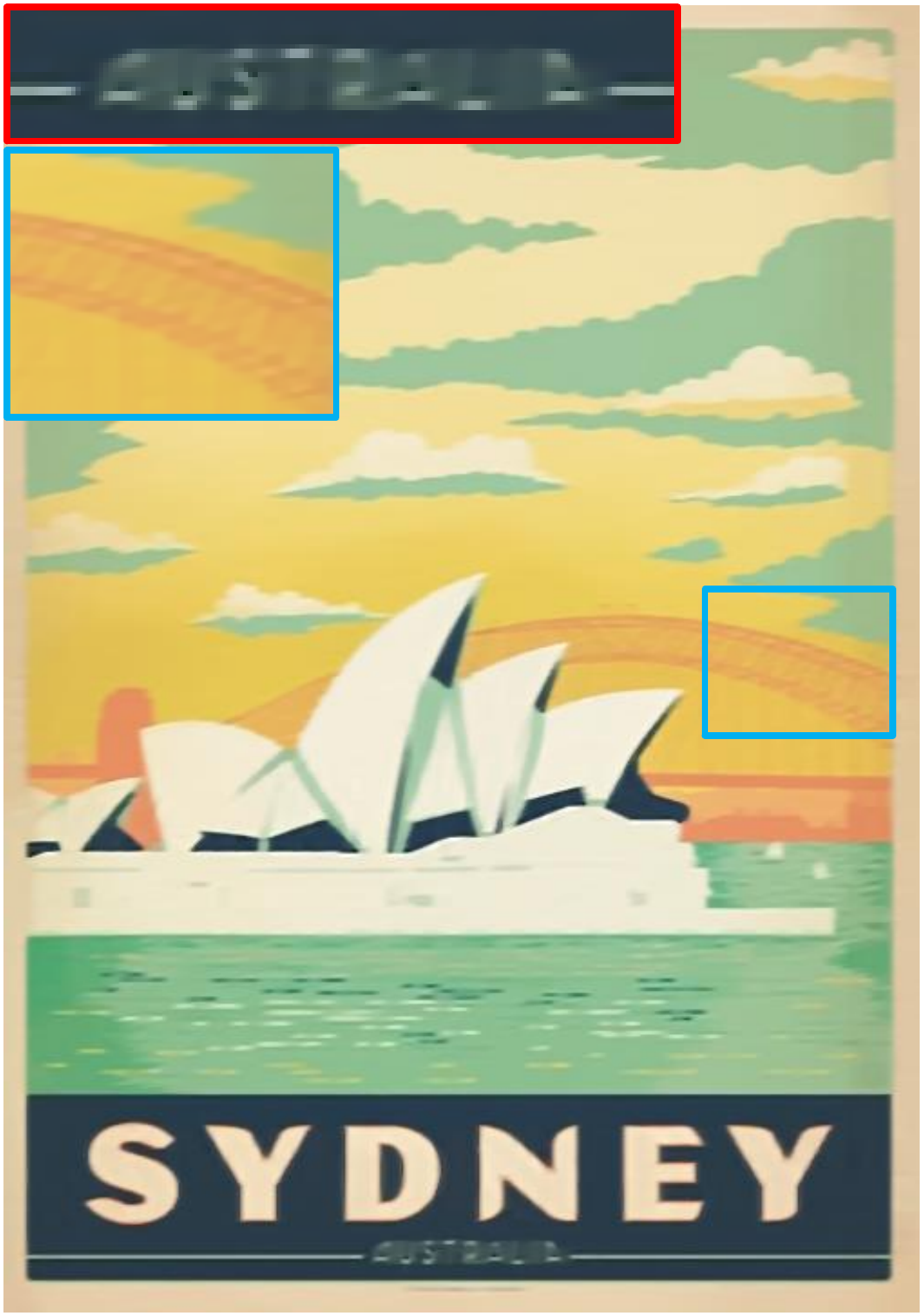}
&\includegraphics[width=0.213\textwidth]{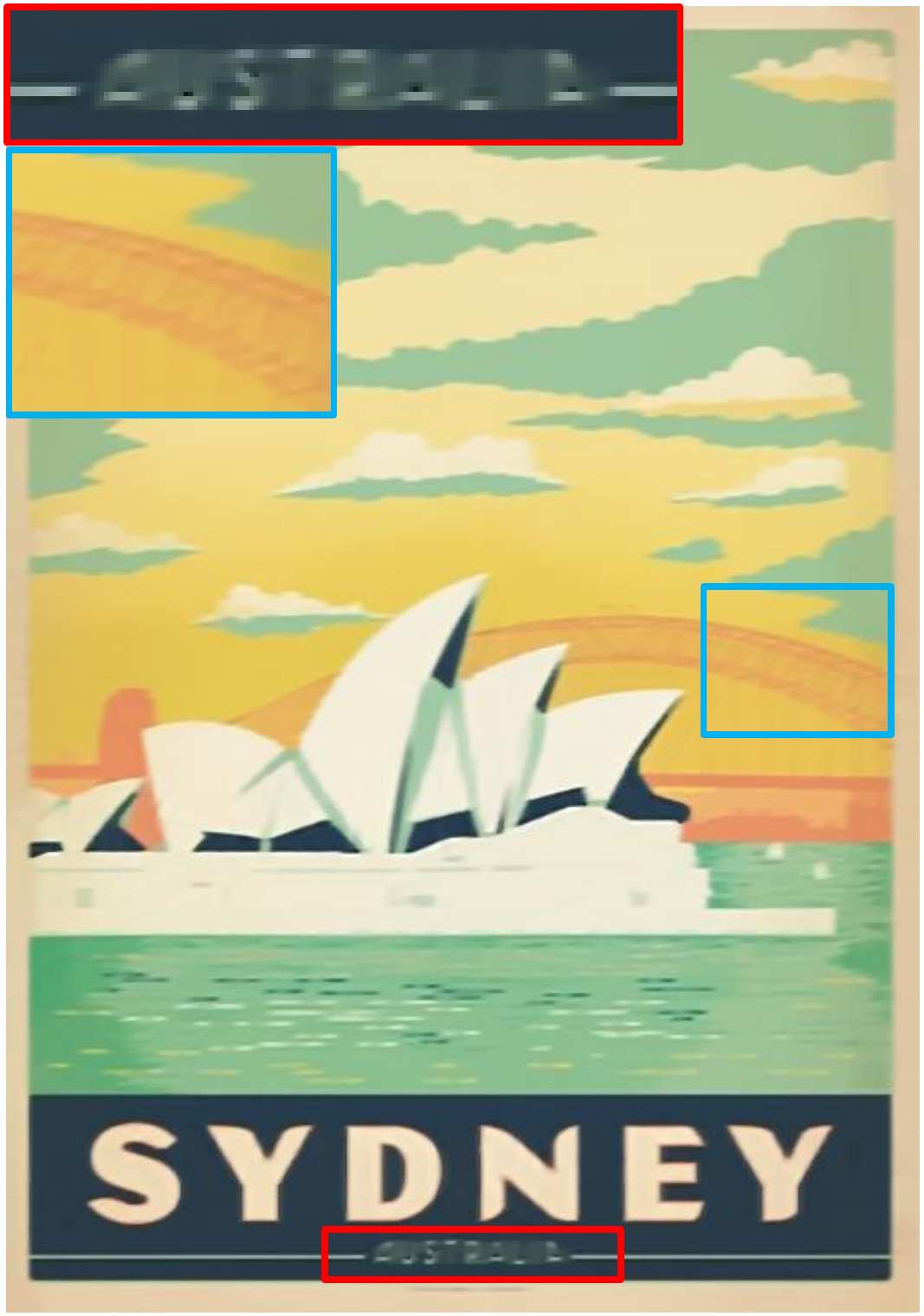}
&\includegraphics[width=0.213\textwidth]{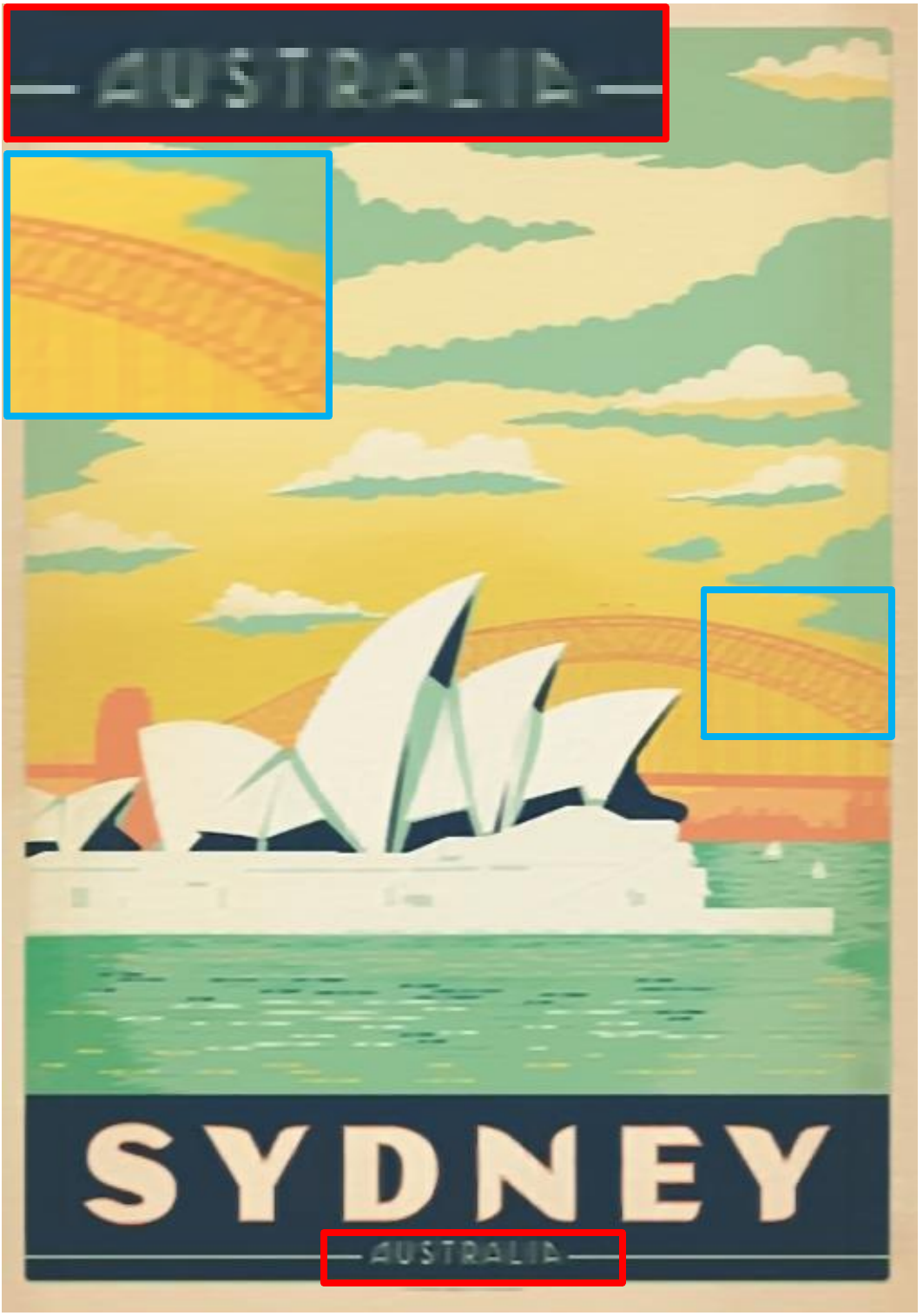}\\
\hspace{-0.25 cm}
(f) Tao \cite{Tao_2018_CVPR} 
& (g) Pan \cite{pan2016blind} 
& (h) Yan \cite{yan2017image} 
& (i) Ours\\
\end{tabular}
\end{center}
\vspace{-3 mm}
\caption{\em \label{fig:fig1}  Our deblurring result compared with the state-of-the-art methods.
(a) Input blurry image. 
(b) The \emph{phase-only image}. 
(c) The auto-correlation for the~\emph{phase-only image}. 
(d) The estimated blur kernel. 
(e) Deblurring result of~\cite{Nah_2017_CVPR}.
(f) Deblurring result of \cite{Tao_2018_CVPR}. 
(g) Deblurring result of \cite{pan2016blind}. 
(h) Deblurring result of~\cite{yan2017image}. 
(i) Our deblurring result. 
(Best viewed on screen). 
}
\end{figure*}
Blind image deblurring aims at estimating the blur kernel and the latent image from an input blurry image. This is an ill-posed problem as there are infinitely many pairs of blur kernels and images that could generate the same blurry image. Blind image deblurring has been extensively studied in computer vision and is still a very active research area~\cite{hyun2015generalized,sellent2016stereo,gong2017motion,Pan_2017_CVPR,Nah_2017_CVPR,Tao_2018_CVPR}, where blur kernel estimation is essentially important in obtaining a high quality sharp image.

Existing blind image deblurring methods tend to formulate the problem within the Maximum A Posteriori (MAP) framework, where the blur kernel and the latent sharp image are optimized jointly. To resolve the ill-posed underlining optimization problem, various assumptions, or regularizations, have been proposed for the blur kernel and the desired latent image, such as the dark channel prior~\cite{pan2016blind}, extreme channel prior~\cite{yan2017image}, $l_0$ regularized prior~\cite{pan2014deblurring,xu2014inverse}, learned image prior using a CNN~\cite{li2018learning}, uniform blur~\cite{levin2009understanding,xu2013unnatural}, non-uniform blur from multiple homographies~\cite{hu2014joint,pan2016soft}, constant depth \cite{gupta2010single,xu2012depth}, in-plane rotation \cite{sun2015learning}, and forward motion \cite{Zheng_2013_ICCV}.
The resultant optimization problem is non-convex in general. The blur kernel and the latent image are usually solved in an alternating fashion. Thus, a proper and effective initialization is demanded to achieve a good~\emph{local optimum solution} and makes the algorithm converge quickly.


  
In this paper, we aim at estimating a high-quality blur kernel directly from the input image with motion blur by studying the problem in the frequency domain. We exploit the \emph{phase-only image} of the input blurry image, which is reconstructed from the Fourier transformed image using the phase information only. The~\emph{phase-only image} contains edge and texture information about the image structure~\cite{oppenheim1981importance,papari2011edge}. The motion (either camera or object motion) information is encoded as repeated image edges in the~\emph{phase-only image} (see Fig.~\ref{fig:fig1} for an example). We show that the auto-correlation of the absolute~\emph{phase-only image} reveals the motion information
including the motion direction and motion magnitude, which is referred to as the \emph{motion pattern} in this paper. It provides information about the blur kernel, thereby leading to a new approach to estimating
the blur kernel.



We further improve the blur kernel and latent image estimation by enforcing a spatial sparsity prior on the kernel as well as the latent image gradient in a simple optimization framework. Furthermore, our blur kernel estimation approach can be naturally extended to handle non-uniform blur in order to deal with the spatially-variant blur kernels that arise in complex image deblurring problems.
Extensive experiment on both synthetic and real images demonstrate the superiority of our approach over the state-of-the-art methods. 





Our main contributions are summarized as follows
\vspace{-2mm}
\begin{enumerate}[1)]
\item We propose a new \emph{phase-only image}-based approach to directly estimating the blur kernel from the input blurry image. The approach for \emph{motion pattern} estimation is easy and efficient, consisting of a few lines of code.
\vspace{-1.5mm}
\item Our single-image blind deblurring model can be naturally extended to handle non-uniform blur in an effective manner. Furthermore, the estimated blur kernel can be easily refined by only enforcing spatial~sparsity.
\vspace{-1.5mm}
\item Evaluated on both synthetic and real images, our proposed approach shows impressive results compared to other state-of-the-art blind deblurring approaches.
\end{enumerate}



\section{Related Work}
\vspace{-1 mm}
\noindent {\bf Single-image blind deblurring.} Single-image deblurring jointly estimates the blur kernel and the latent sharp image from the blurry one, which is highly under-constrained since the blurry image could be explained by many pairs of blur kernel and sharp image \cite{ji2008motion,pan2019single}. In general, image deblurring is formulated in a MAP framework with~\emph{priors} on blur kernels or latent images. The Sparsity prior has proved effective in blur kernel estimation. For instance, Krishnan \etal \cite{krishnan2011blind} applied normalized sparsity in their MAP framework to estimate the blur kernel.
Xu \etal \cite{xu2013unnatural} proposed an approximation of
the $l_0$-norm as a sparsity prior in order to jointly estimate
sharp image and blur kernels. Edge-based methods for blur kernel estimation have
been exploited recently \cite{xu2010two,joshi2008psf,cho2009fast,sun2013edge}. Xu~\etal~\cite{xu2010two} proposed a two-phase method for single-image deblurring. The blur kernel is first estimated based on the selected image edges and refined by ISD optimization.  The latent sharp image is then restored by total-variation (TV)-$l_1$ deconvolution. In addition, a Gaussian prior is imposed to help the estimation of the blur kernel~\cite{joshi2008psf,cho2009fast}, which leads to an efficient solver.
Moreover, the blur kernel has been modelled based on various motion assumptions, such as in-plane camera rotation \cite{sun2015learning} or camera forward motion \cite{Zheng_2013_ICCV}. A few works have exploited the layer-wise scene structure to model the blur kernel~\cite{gupta2010single,hu2014joint,pan2016soft}. 
Gupta \etal \cite{gupta2010single} represent the camera motion trajectory using a motion density function, which requires a constant depth or fronto-parallel scene assumption. Hu \etal \cite{hu2014joint} proposed jointly estimating the depth layering and remove the blur caused by in-plane motion from a single blurry image. Pan \etal \cite{pan2016soft} proposed jointly estimating object segmentation and camera motion by incorporating soft segmentation. Note that both approaches require user input for initial depth layer segmentation. 
 
\vspace{0.5mm}
\noindent{\bf{Video image blind deblurring.}}
In order to better model non-uniform blur, monocular video and stereo based deblurring approaches are proposed to handle blurring in realistic scenes \cite{pan2018depth,xu2012depth}.
Cho \etal \cite{cho2012video} proposed a method relying on the assumption that
salient sharp frames frequently exist in videos, which only allows for slowly moving objects in dynamic scenes. Wulff and Black \cite{wulff2014modeling} proposed a layered model to estimate both foreground motion and background motion. However, these motions are restricted to affine models, and it is difficult to extended them to multi-layer scenes due to the difficulty in depth ordering. 
Kim and Lee \cite{kim2014segmentation} incorporated optical flow estimation to guide the blur kernel estimation, which is able to deal with certain object motion blur. In \cite{hyun2015generalized}, a new method is proposed to simultaneously estimate optical flow and tackle general blur by minimizing a single non-convex energy function. 
Stereo images and videos can provide depth information which allows to better model pixel-wise blur kernel. 
Sellent \etal \cite{sellent2016stereo} proposed a stereo video deblurring technique, where 3D scene flow is estimated from the blurry images using a piecewise rigid scene representation. Pan \etal \cite{Pan_2017_CVPR} proposed a single framework to jointly estimate the scene flow and deblur the images. 

\vspace{0.5mm}
\noindent{\bf Deep learning based image deblurring.} 
Recently, image deblurring has greatly benefited from the great advances in deep learning \cite{DeblurGAN,sun2015learning,Zhang_2018_CVPR,Tao_2018_CVPR}.
Sun \etal \cite{sun2015learning} proposed a convolutional neural network (CNN) to estimate locally linear blur kernels. 
Gong \etal \cite{gong2017motion} learned optical flow field from a single blurry image directly through a fully-convolutional deep neural network. The blur kernel is then obtained from the estimated optical flow which is applied in an MAP framework to restore the sharp image.
Su \etal \cite{Su_2017_CVPR} trained an end-to-end CNN to accumulate
information across frames for video deblurring. Nah \etal~\cite{Nah_2017_CVPR} proposed a multi-scale
CNN that restores latent images in an end-to-end learning manner without
any assumption on the blur kernel model. Li \etal \cite{li2018learning} used a
learned image prior to distinguish whether an image is sharp or not and
embedded the learned prior into the MAP framework. Tao \etal \cite{Tao_2018_CVPR} proposed a light and compact network, SRN-DeblurNet, to deblur the image.
While achieving reasonable performance on various scenarios, the success of these
deep learning based methods depends on the consistency between the training
datasets and the testing datasets, which can hinder the generalization
ability.

\begin{figure}[t]
\begin{center}
\begin{tabular}{cc}
\includegraphics[width=0.186\textwidth]{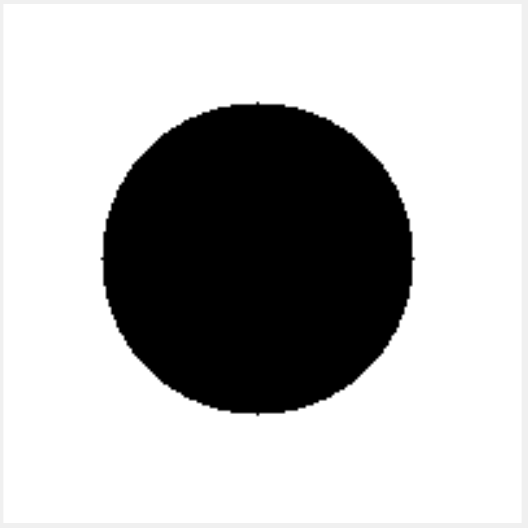}
&\includegraphics[width=0.186\textwidth]{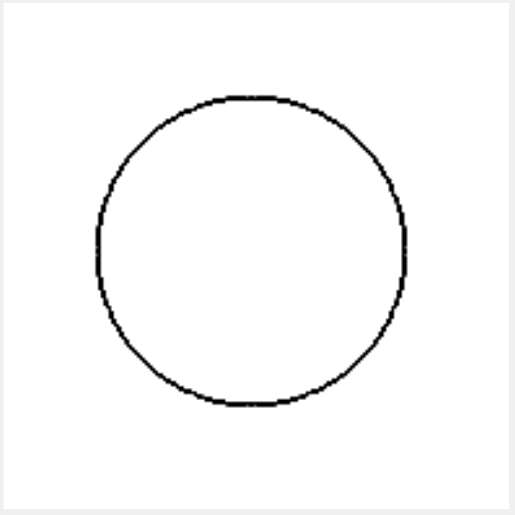}\\
(a) Sharp Image  
& (b) $|P(\vL)|$\\
\includegraphics[width=0.186\textwidth]{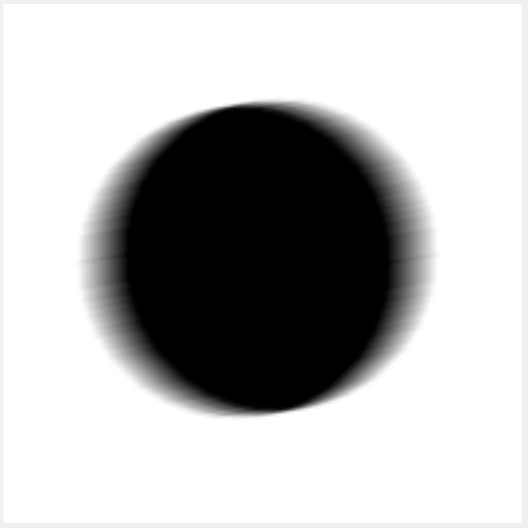}
&\includegraphics[width=0.186\textwidth]{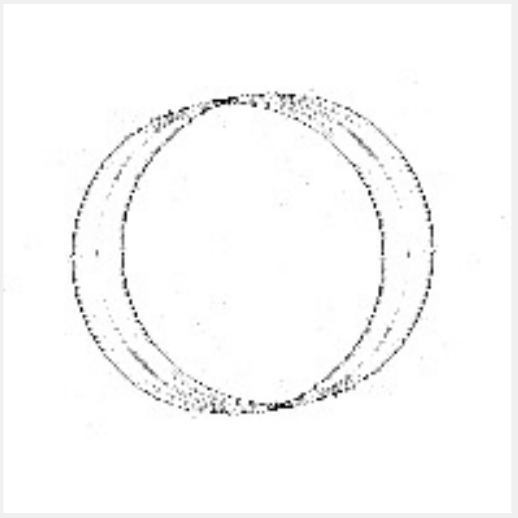}\\
(c) Blurry Image
& (d) $|P(\vB)|$\\
\end{tabular}
\end{center}
\vspace{-1.4mm}
\caption{\em \label{fig:circle} We use a circle image as an example. The image is blurred by a linear kernel, where the kernel length is 20 pixels and the direction is 10 degree. 
}
\end{figure}
\section{Method}

\begin{figure*}[t]
\vspace{-2mm}
\centerline{
\begin{tabular}{cc}
\includegraphics[height=0.2\textwidth]{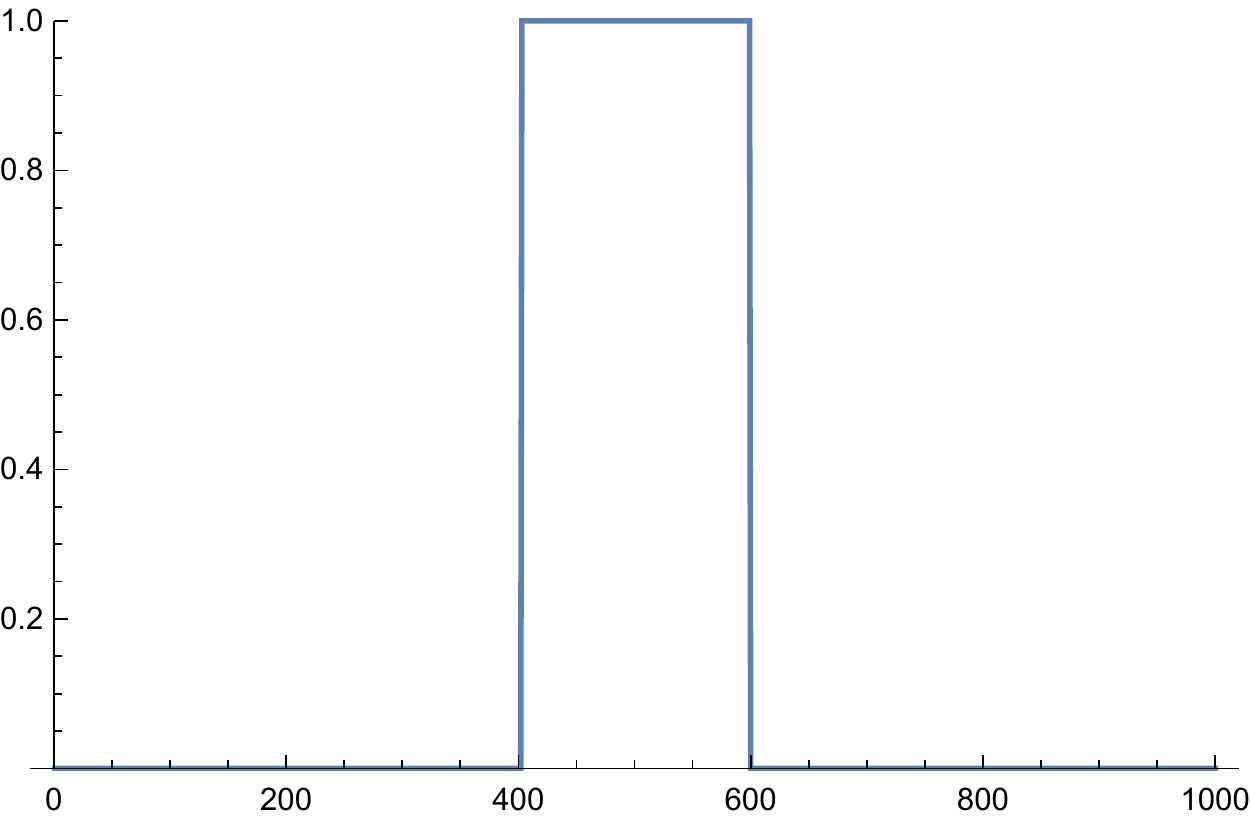} 
\hspace{2 cm}
& \includegraphics[height=0.2\textwidth]{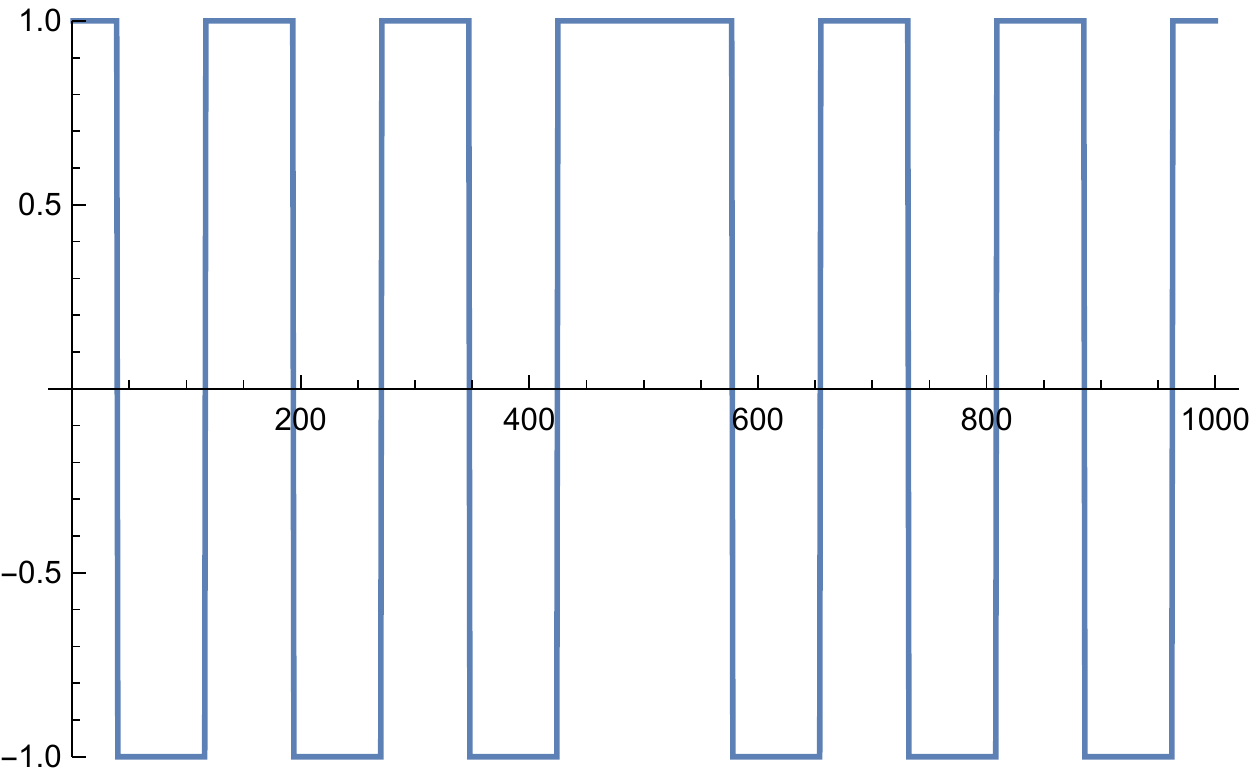}\\
(a) 
\hspace{2 cm}
&(c)\\
\includegraphics[height=0.20\textwidth]{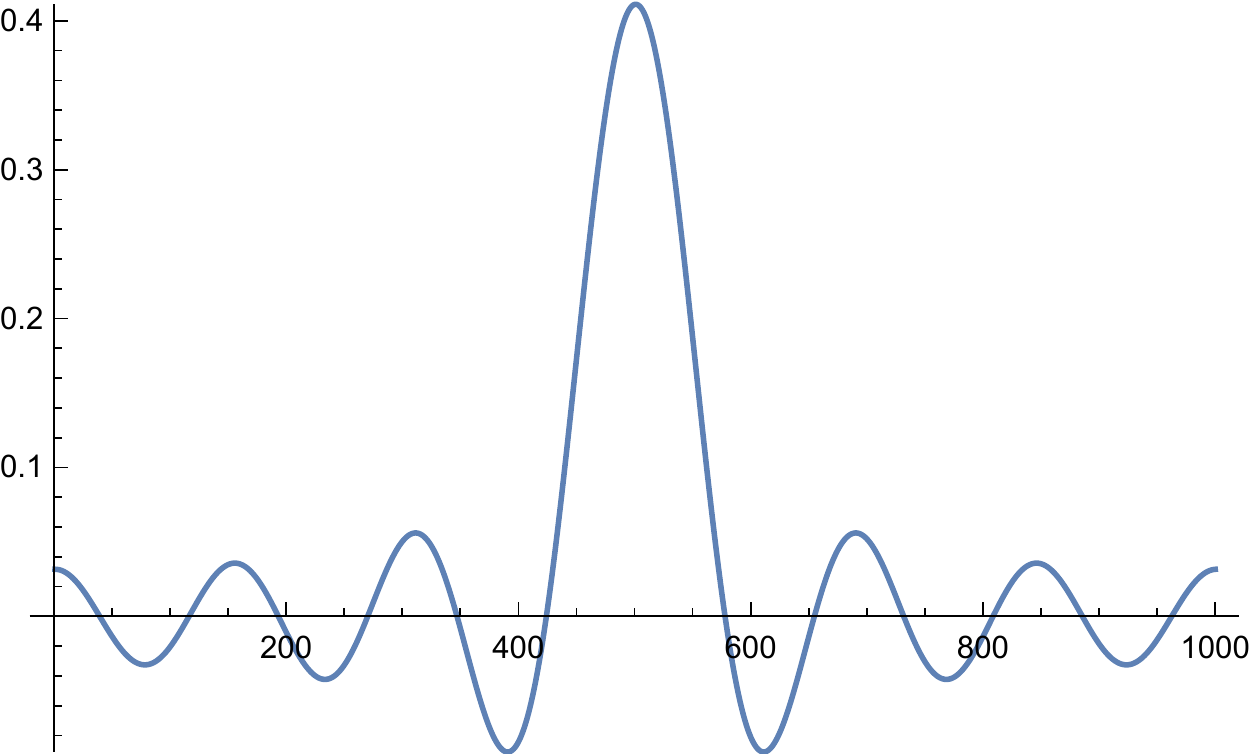}
\hspace{2 cm}
&\includegraphics[height=0.2\textwidth]{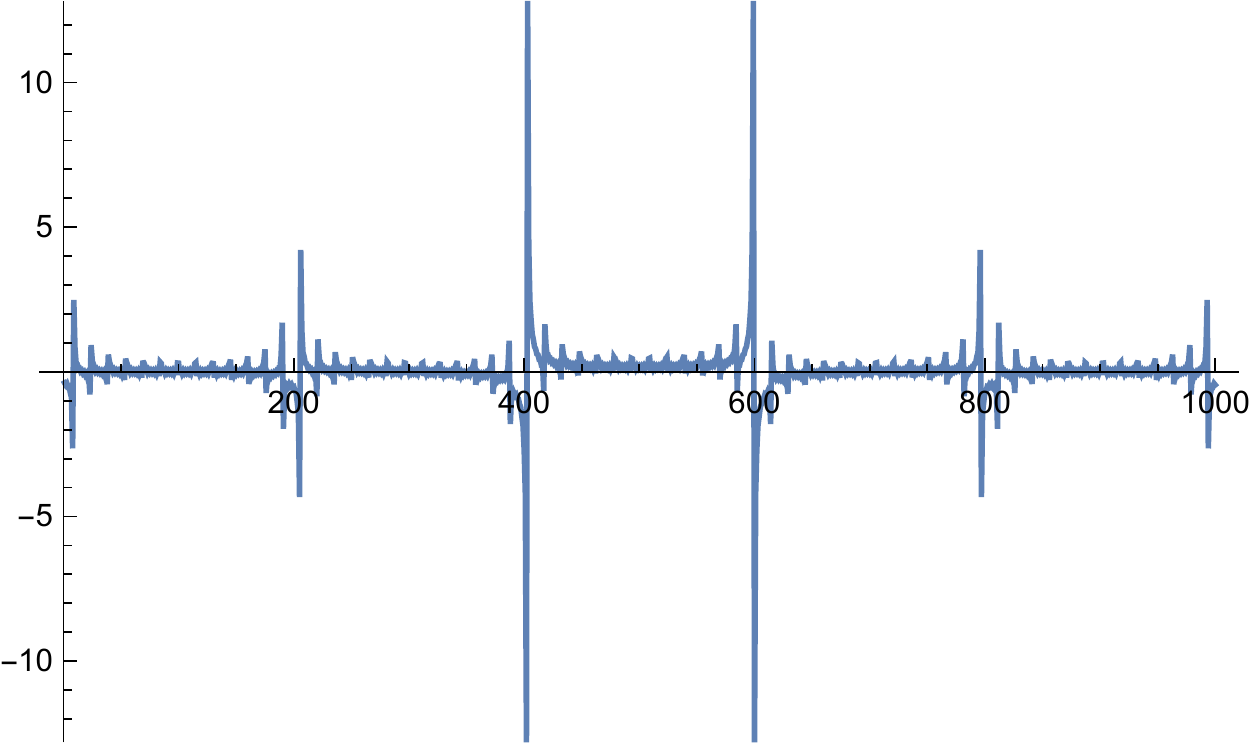}\\
(b) 
\hspace{2 cm}
& (d)
\end{tabular}
}
\vspace{0.5 mm}
\caption{\em Given a top-hat function (a), its fourier transform is a $sinc$ shown in (b). (The central peak has twice the width of the others. Note that since the top-hat is symmetric, its Fourier transform is real, hence its phase is either $+1$ or $-1$ shown in (c).) The phase-only image of the top-hat shown in (d) is obtained by taking the inverse Fourier transform of the function in (c).}
\label{fig:fourierTheory}
\end{figure*}

\subsection{Fourier Theory of Phase-only Images}
\label{sec:deblurring-theory}

This section contains the main theoretical insights of this paper.
Our goal is to find the latent sharp image from a single blurry image. The
blurry image can be modelled as a convolution of the latent image with a blur
kernel,
\begin{equation}\label{eq:blurmodel}
\vB = \vL\otimes\vk,
\end{equation}
where $\vB$ is the known blurry image,
$\vL$ denotes the latent sharp image,
$\vk$ is the blur kernel, $\otimes$ is the convolution operator. Note that this problem is highly under-determined since multiple pairs of $\vL$ and $\vk$ can lead to the same blurry image. 

In the Fourier domain, Eq.~(\ref{eq:blurmodel}) corresponds to $\FT(\vB) = \FT(\vL) \odot \FT(\vk)$, where $\odot$ represents the
component-wise multiplication.

The phase and amplitude of a complex number
$z = k e^{i\theta}$ are $e^{i\theta}$ and $k\ge 0$ respectively.  Applying these
component-by-component to a Fourier transformed image $\FT(\vL)$ gives the
phase and amplitude components.  We denote taking the
phase of a complex signal by $\calP(\cdot)$.  Taking the inverse Fourier
transform of the phase-component gives the 
{\em phase-only image}, $P(\vL) = \FT^{-1}(\calP (\FT(\vL)))$.  
It is well known that the phase-only image bears more similarity
to the original image than the analogously defined amplitude image.
Fig.~\ref{fig:circle} shows an example of the phase-only image derived from
a clean and blurry image. 
As may be observed, 
taking a phase-only image acts as a sort of edge-extractor.  
This is related to the fact, noted in \cite{kovesi2003phase} 
that the Fourier components of an edge tend to be in-phase with each other.
For a real image $\vL$, the phase-only image will also be real.
Another simple property is {\em rotation-covariance}: if $R$ represents
rotation then $P(R(\vL)) = R(P(\vL))$. It is also shift-covariant.

We now make a basic observation regarding the phase-only image
of a convolution.
\vspace{0.5mm}
\begin{lemma}
\label{lem:phase-blur}
The phase-only image of a convolution 
$P(\vL \otimes \vk)$, equals the convolution of the 
phase-only image and the phase-only kernel.
\begin{align}\label{eq:phase-blur}
P(\vL \otimes \vk) &= \FT^{-1} (\calP (\FT (\vL \otimes \vk))) 
= P(\vL) \otimes P(\vk) ~.
\end{align}
\end{lemma}
This results from a simple calculation.

\vspace{-2.5 mm}
\paragraph{Linearly-blurred image. }


For a simple linear (straight-line) blur kernel, the form of $P(\vk)$
can be computed. 
By rotation and shift covariance, it may be assumed without loss of 
generality, that $\vk$
is axis-aligned, in which case $\vk(x, y) = \delta(y) H(x)$,
where $\delta(y)$ is a Dirac delta function and $H(x)$ is a top-hat.
The Fourier transform is separable, so it follows that
$P(\vk)(x, y) = \delta(y) P(H)(x)$.
Hence, we investigate what the $1D$
phase-only signal $P(H)$ is.
The result is shown in 
Fig.~\ref{fig:fourierTheory}.
A formula for the shape of the phase-only top-hat of width $w$
is derived (for the continuous Fourier Transform) in the supplementary material, and is 
equal to $(\sqrt{2\pi}/w)\,\sinc(\pi x/w) /\cos(\pi x/w)$,
which is plotted in 
Fig.~\ref{fig:fourierTheory}(d).
More details of the properties of this function are
given in the supplementary material.

\begin{figure*}[!htb]
\begin{center}
\begin{tabular}{cccc}
\hspace{-0.25 cm}
\includegraphics[width=0.210\textwidth]{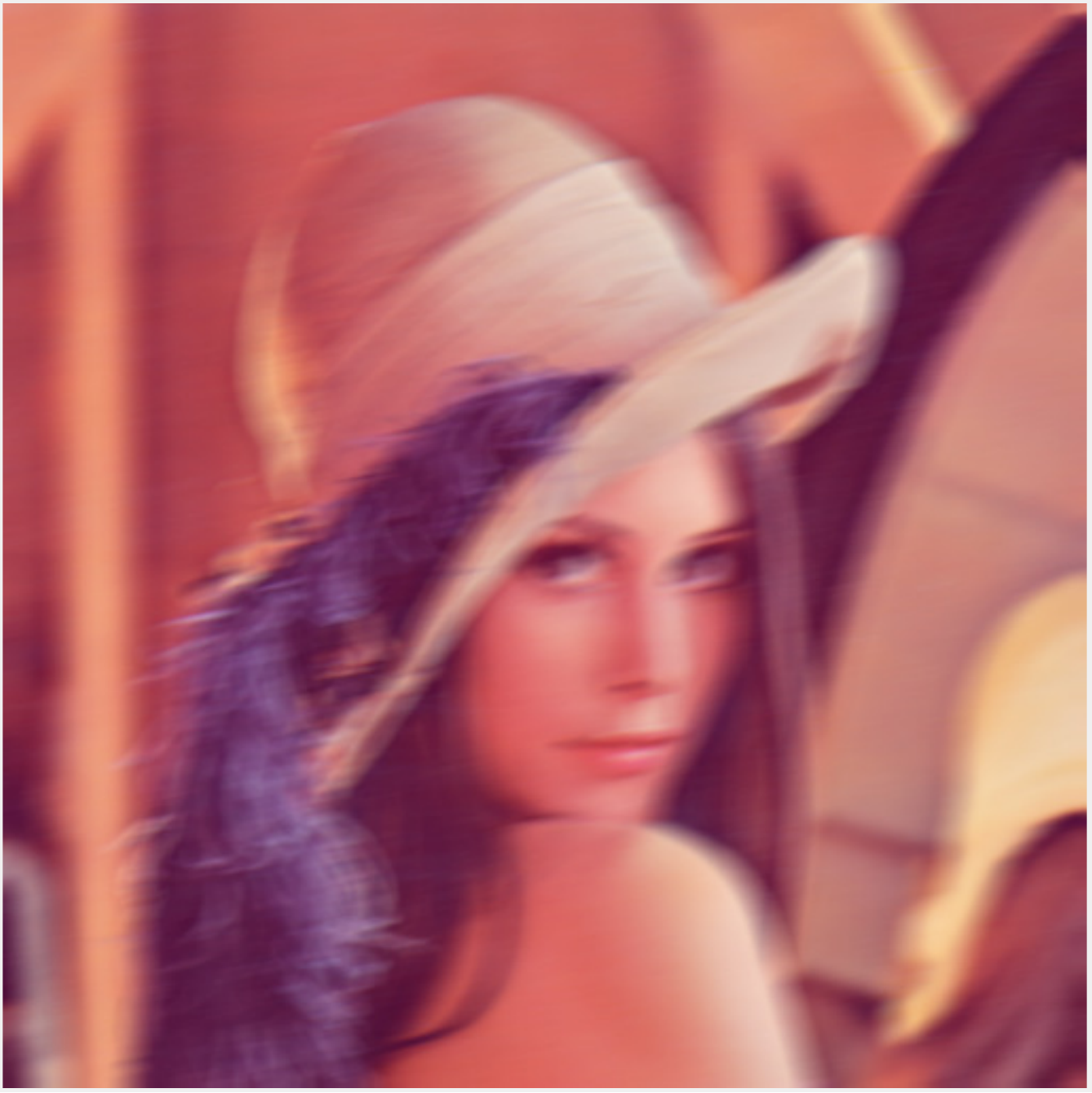}
&\includegraphics[width=0.210\textwidth]{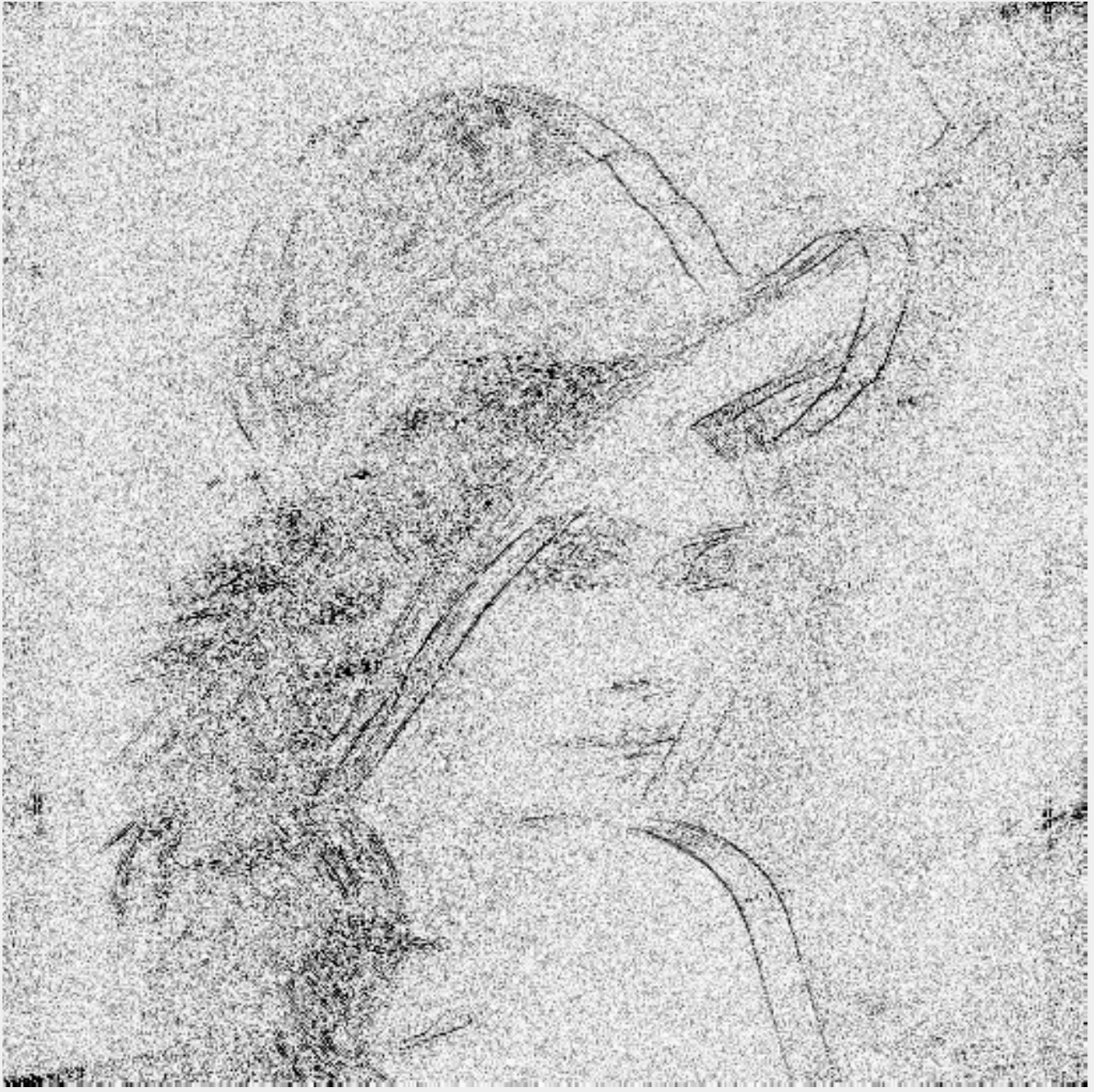}
&\includegraphics[width=0.210\textwidth]{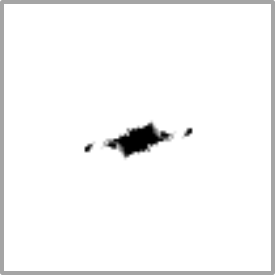}
&\includegraphics[width=0.210\textwidth]{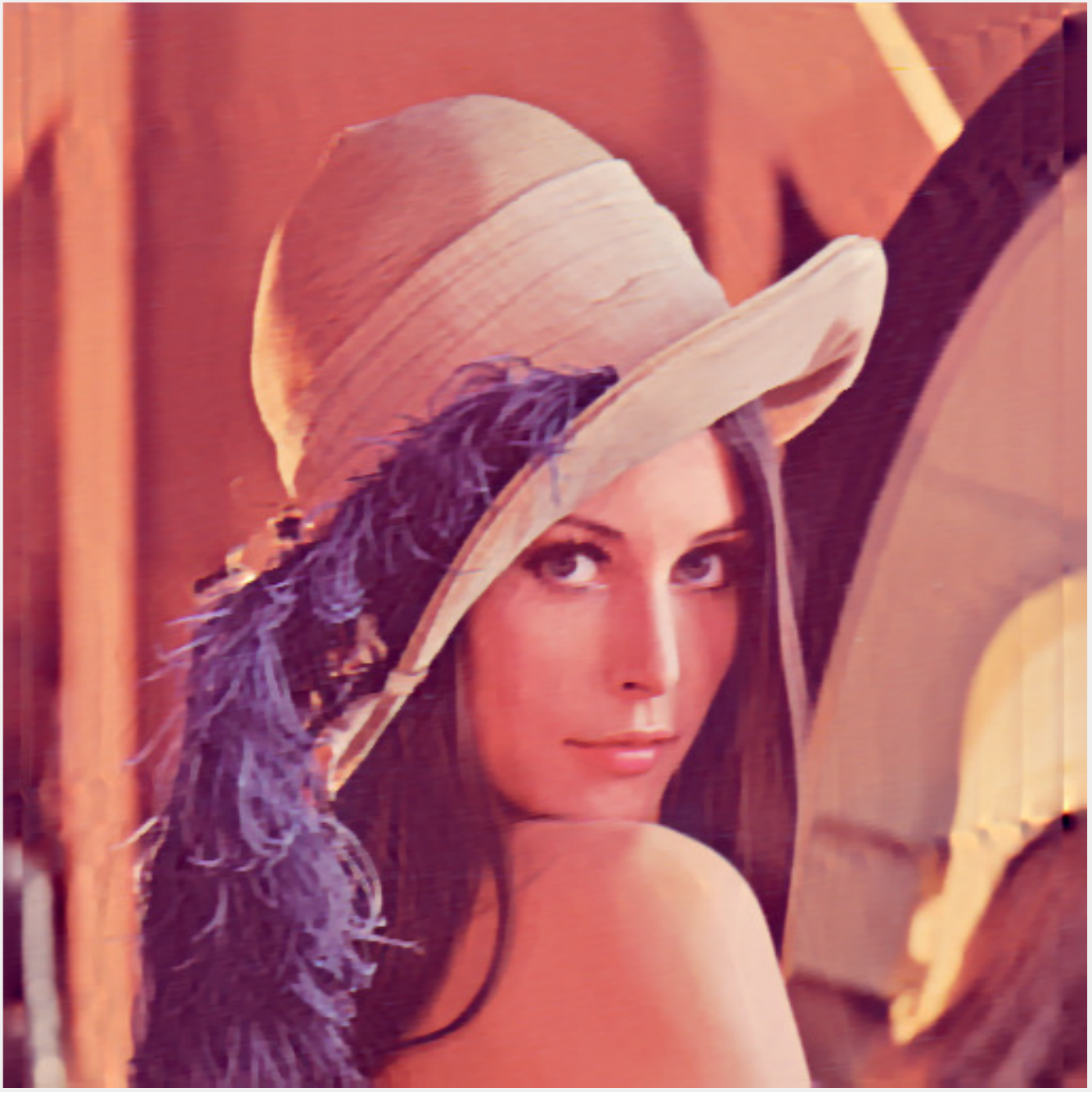}\\
\hspace{-0.25 cm}
\includegraphics[width=0.210\textwidth]{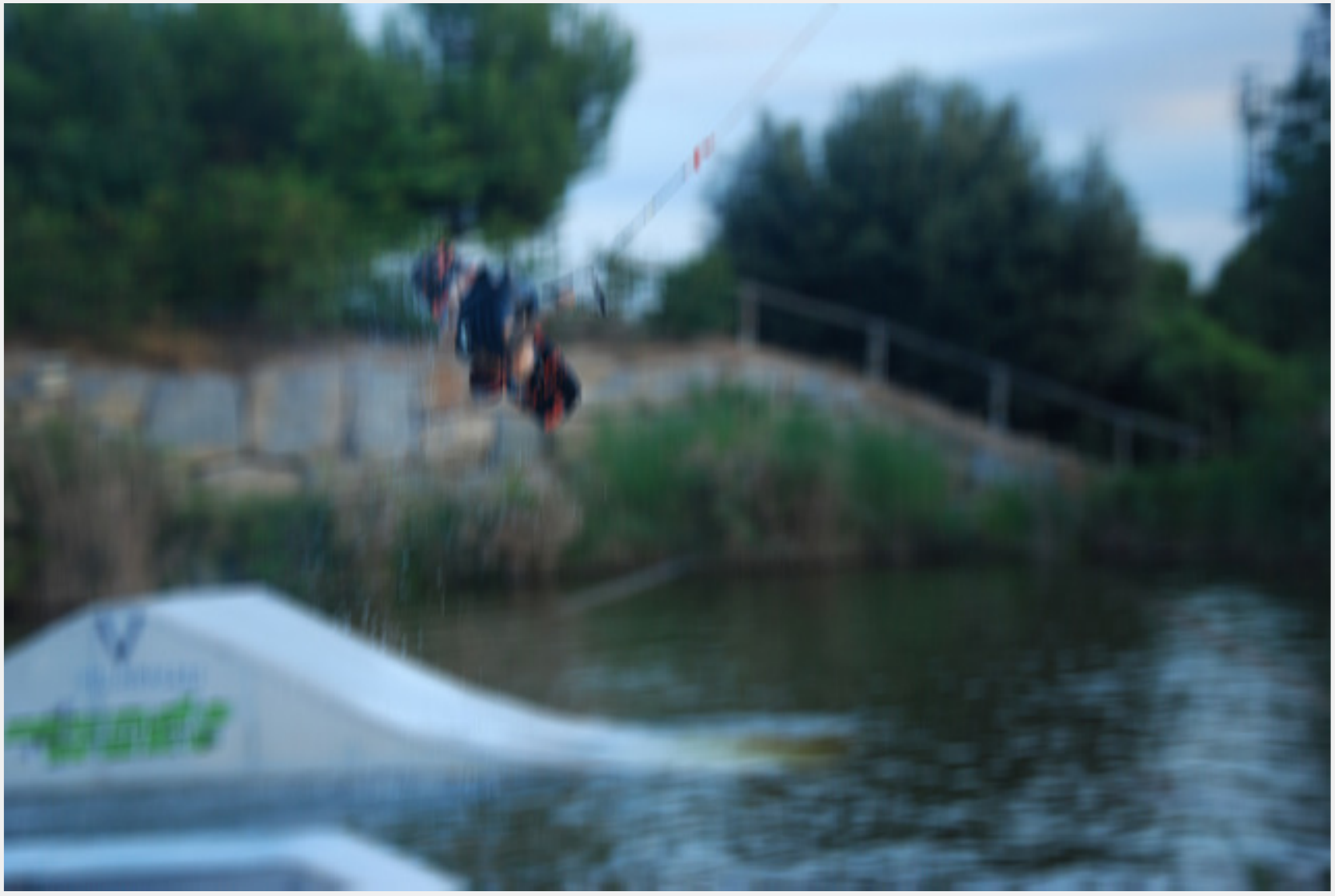}
&\includegraphics[width=0.210\textwidth]{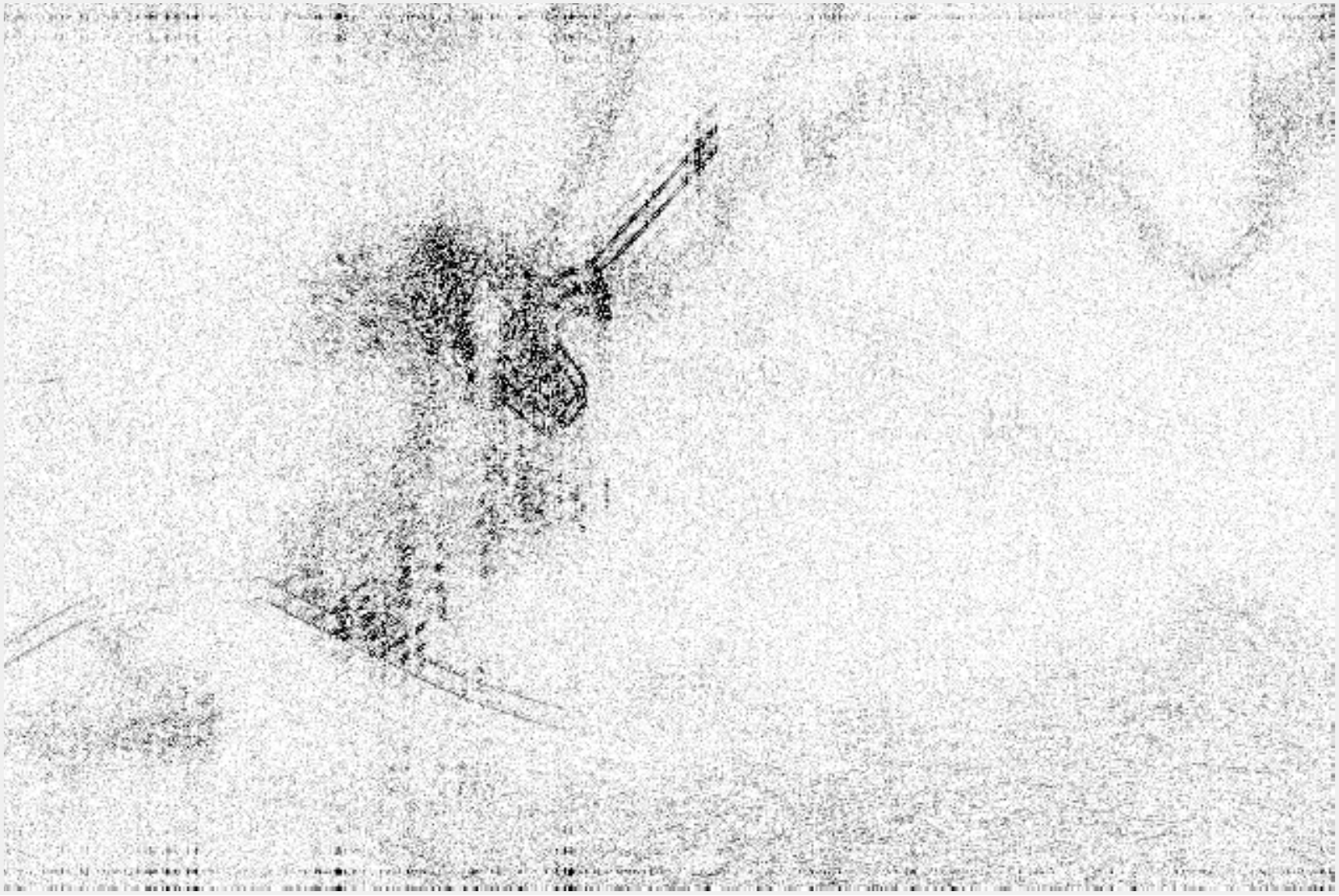}
&\includegraphics[width=0.210\textwidth]{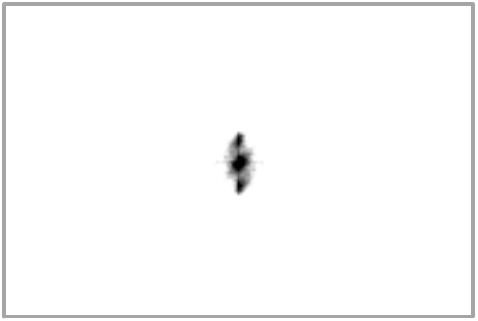}
&\includegraphics[width=0.210\textwidth]{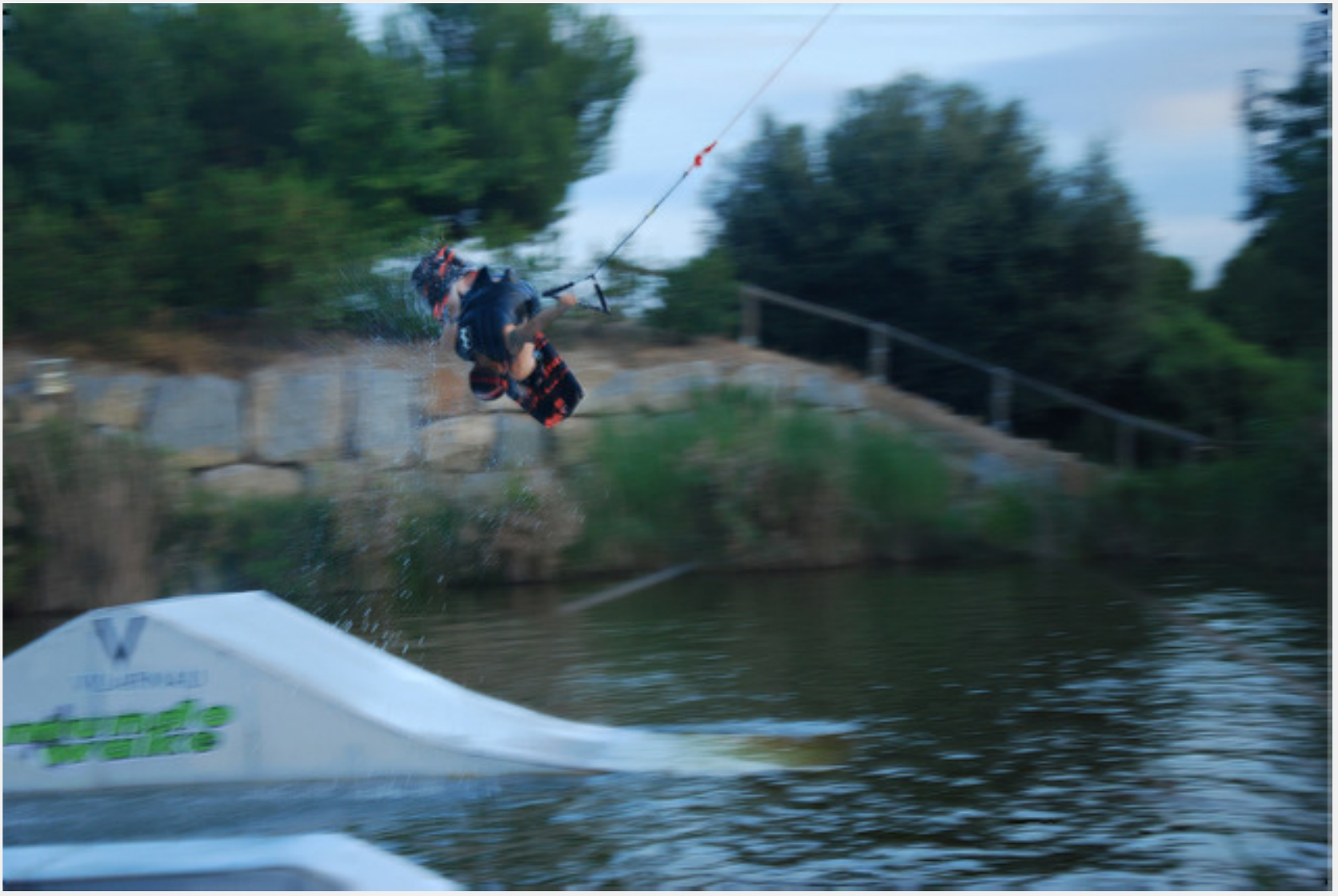}\\
\hspace{-0.25 cm}
 (a) Blurry Image  
 & (b) $|P(\vB)|$
& (c) $\calA(|P(\vB)|)$ 
& (d) Deblurring Results\\
\end{tabular}
\end{center}
\vspace{-2 mm}
\caption{\em \label{fig:phase_corr} 
(a) Input blurry images, the top one is a synthetic image created by ourselves and the bottom one is a real image from dataset \cite{shi2014discriminative}. 
(b) The absolute phase-only image of the blurry image, $|P(\vB)|$,
results in two principal copies (others more faint) of 
$P(\vL)$.
(c) The autocorrelation of the absolute phase-only image, $\calA(|P(\vB)|)$, showing two distinct peaks (separated by the length of the filter kernel).
Distinguishing the two principal peaks of the autocorrelation 
(apart from the origin) can be used to determine the orientation 
and width of a linear (straight-line) blur kernel. (d) shows our deblurring results with sharp edges.}
\end{figure*}

According to 
Eq.~(\ref{eq:phase-blur}), if $\vB = \vL \otimes \vk$, then 
$P(\vB)$ is obtained by convolving $P(\vL)$ in the orientation of
the linear kernel with the phase-only kernel, shown in 
Fig.~\ref{fig:fourierTheory}(d).
This results in the creation of multiple copies (``ghosts''),
of the phase-only image, $P(\vL)$, separated by
the width of the filter.  (The copies due to the principal peaks
will be the most noticeable.)%
\footnote{
A more exact statement is that $P(\vB)$ consists of multiple ghosts,
separated by the filter width, of the {\bf gradient} of $P(\vL)$
in the filter direction.  An exact
derivation is given in the supplementary material. This includes
also an exact derivation of $P(H)$.
}
This is shown in \fig{phase_corr}.

\vspace{-2 mm}
\paragraph{The key advantage of phase-only image. }
This analysis and the examples show the advantage and purpose 
in considering the phase-only image as a means of determining the blur
kernel, and subsequently deblurring the image.  This is illustrated by the
analysis of the linear kernel.

The effect of blurring is to smear the image in the blur direction,
as shown in Fig.~\ref{fig:phase_corr} (top left).  From this image,
it is not easy to discern the shape of the kernel, particularly the linear extent
of the kernel.  On the other hand, in the phase-only image, the effect of blurring
is to create {\em two principal identical copies} of $P(\vL)$ separated by the extent of
the blur kernel.  This is immediately evident from Fig.~\ref{fig:phase_corr}(b),
or Fig.~\ref{fig:circle}(d).  Thus, the continuous
smear in the blurred image is replaced by a simple sum of two (principle) copies in the phase-only blurred image.
This simplification of the effect of 
blurring makes the further image-processing to compute the blur-kernel much simpler.

This discovery of the application of the phase-only image to deblurring is the
key original contribution of this paper, and
the supplementary material provides a rigorous mathematical justification of the
empirical observation, which we hope the reader will enjoy.

\subsection{Autocorrelation}
Using phase-only to obtain $P(\vB)$ 
from a blurry image results in multiple (two principal) shifted
copies of $P(\vL)$.  Note that $P(\vL)$ is not known.  However, this
suggests the use of autocorrelation of $P(\vB)$.  

Autocorrelation of a signal $\vI$ ($1$ or $2$-dimensional) 
is computed using Fourier transform as:
%
\[
\calA (\vI) = \FT^{-1} (\FT(\vI) \odot \overline{\FT(\vI)}).
\]
%
Unfortunately, if $\vI$ is itself a phase-only image,
derived from $\vJ$, then
%
\[
\FT(\vI) = \FT(\FT^{-1} \calP(\FT(\vJ)))  = \calP(\FT(\vJ)).
\]
So
{\small
~~\(
\calA(\vI) \!=\!
\FT^{-1} ( \calP(\FT(\vJ))  \odot \overline{\calP(\FT(\vJ))} )
\!=\! \FT^{-1} (1) 
\!= \delta
\)\\
}
where $\delta$ is a Dirac delta function at the origin.
In other words, a phase-only image 
{\bf is completely un-selfcorrelated}.

In other words, we cannot derive any information whatever from
the autocorrelation of a phase-only image.
The solution is to use the absolute value of the phase-only
image instead.  In other words, we compute $\calA(|P(\vB)|)$, which should show the desired behaviour.


Fig.~\ref{fig:phase_corr} shows the absolute~\emph{phase-only image} $|P(\vB)|$ and its autocorrelation $\calA(|P(\vB)|)$. It is noticed that multiple copies of $|P(\vL)|$ are shown in $|P(\vB)|$. 
The most noticeable repeated edges are due to the principal peak of $P(\vk)$ (as analyzed above) indicating the start and end point of the moving camera.

The autocorrelation of the absolute~\emph{phase-only image} shows several bright points that indicate the motion of the camera, \eg, the motion direction and magnitude, which is referred to as~\emph{motion pattern}.  The autocorrelation image will consist of a central peak 
plus two side-peaks separated by the extent (and in the direction) of the 
blur-kernel.

Consequently, the motion of the camera will provide faithful information for obtaining the blur kernel. Therefore, in the following section, we will present our approach to image deblurring based on the analysis of the autocorrelation of the absolute~\emph{phase-only image}.
\section{Uniform Deblurring}
\label{sec:uniformDeblurring}


Based on the analysis of the Fourier theory of phase-only images, we introduce our approach to estimate the blur kernel and deblur the images.


\subsection{Uniform Blur from Linear Motion}
Consider the blur caused by a pure linear motion. By computing the autocorrelation of the absolute~\emph{phase-only image}, the~\emph{motion pattern}, namely the motion direction and the motion magnitude, is extracted by directly connecting the two end bright points in $\calA(|P(\vB)|)$. The blur kernel is then formed based on the extracted~\emph{motion pattern}. In particular, the motion magnitude determines the kernel size. The non-zero kernel values are uniformly distributed along the motion direction (see Fig.~\ref{fig:phase_corr} the top row for an example). Given the built blur kernel, the latent image can be easily obtained by solving the Eq.~(\ref{eq:energy1}) which will be introduced in the following section.

\begin{figure}[t]
\begin{center}
\begin{tabular}{cc}
\hspace{-0.3cm}
\includegraphics[width=0.23\textwidth]{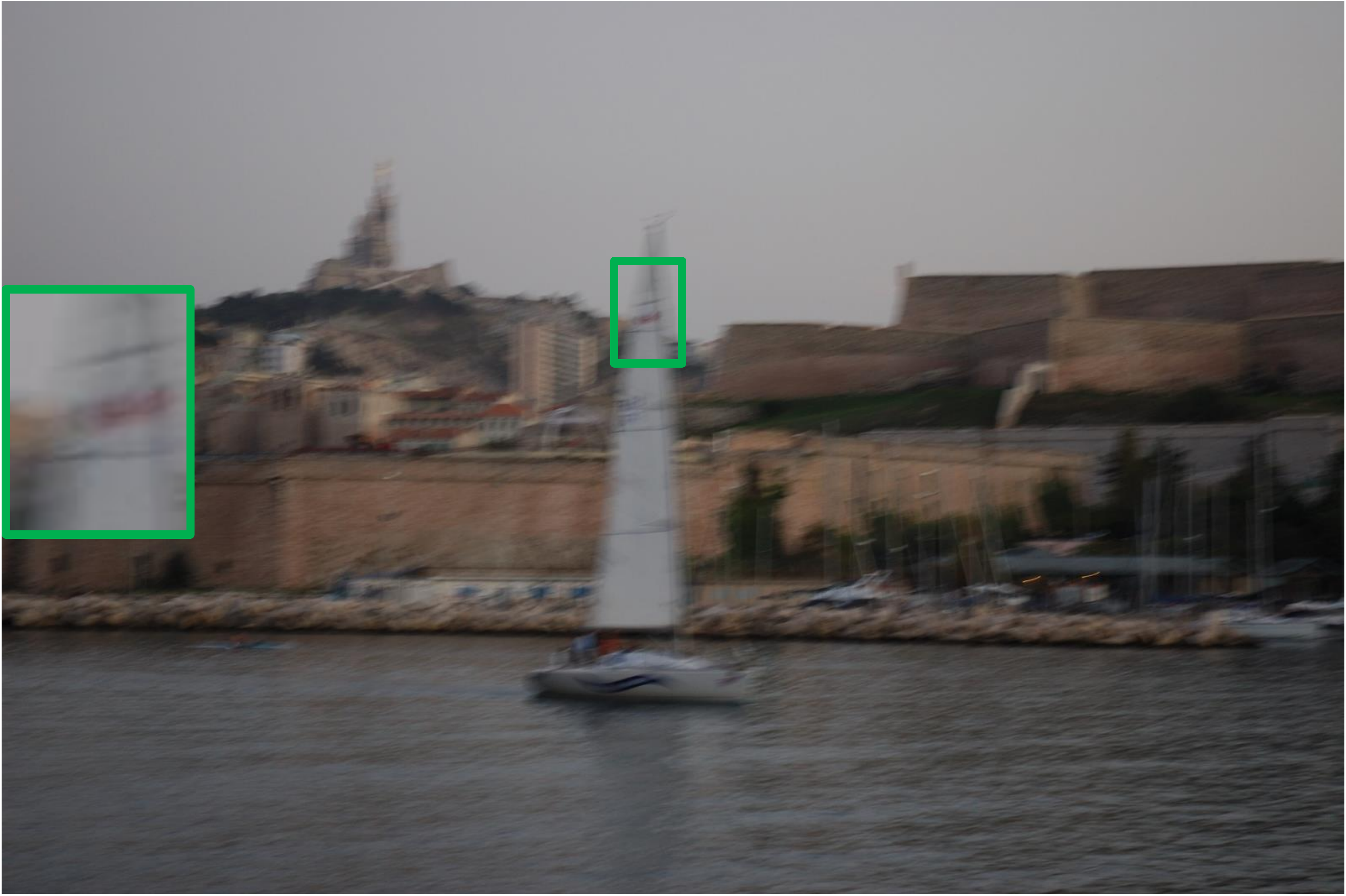}
\hspace{-0.3cm}
&\includegraphics[width=0.23\textwidth]{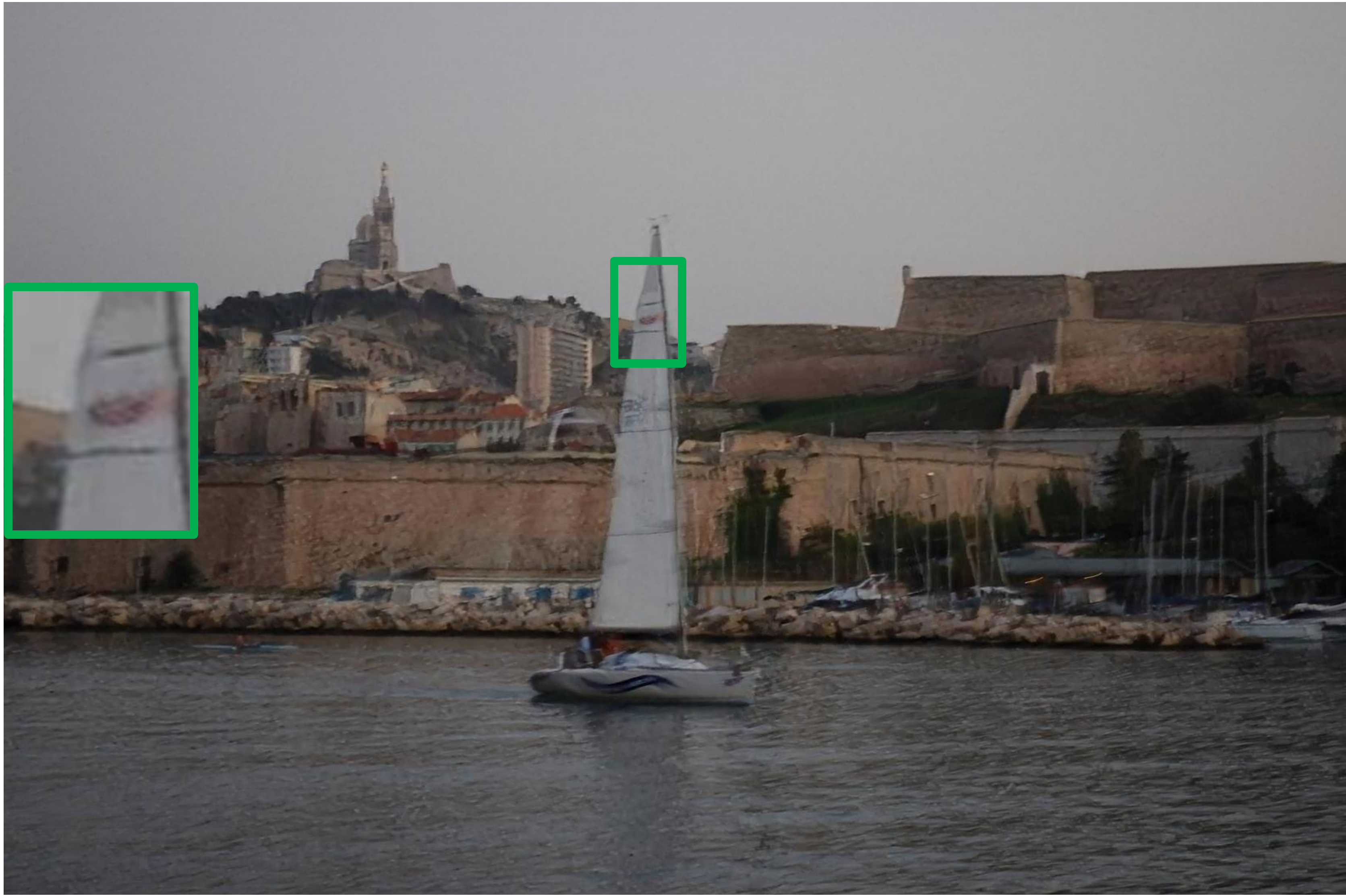}\\
\hspace{-0.3cm}
(a) Blurry Image  
\hspace{-0.3cm}
& (b) Nah \cite{Nah_2017_CVPR} \\
\hspace{-0.3cm}
\includegraphics[width=0.23\textwidth]{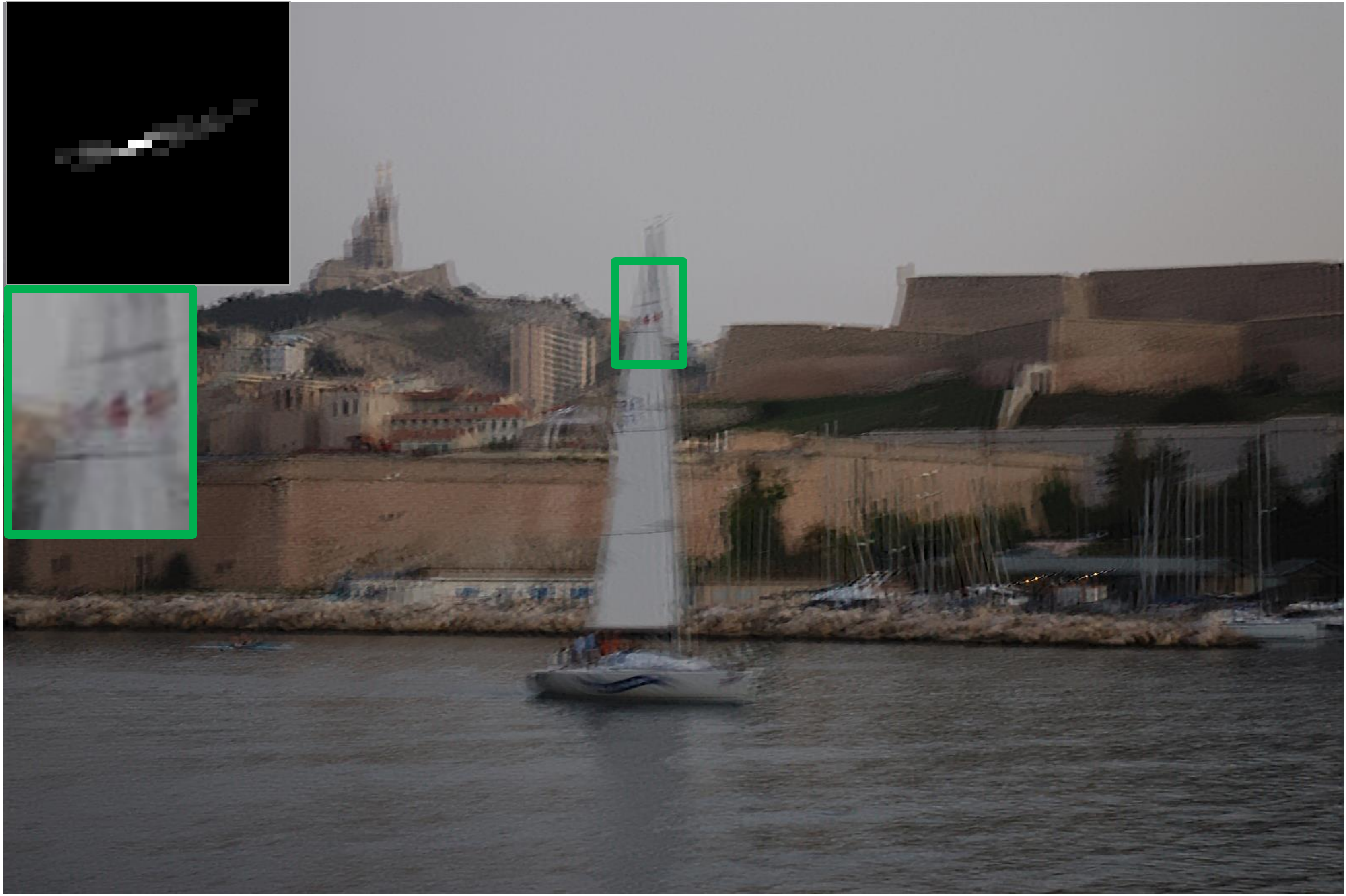}
\hspace{-0.3cm}
&\includegraphics[width=0.23\textwidth]{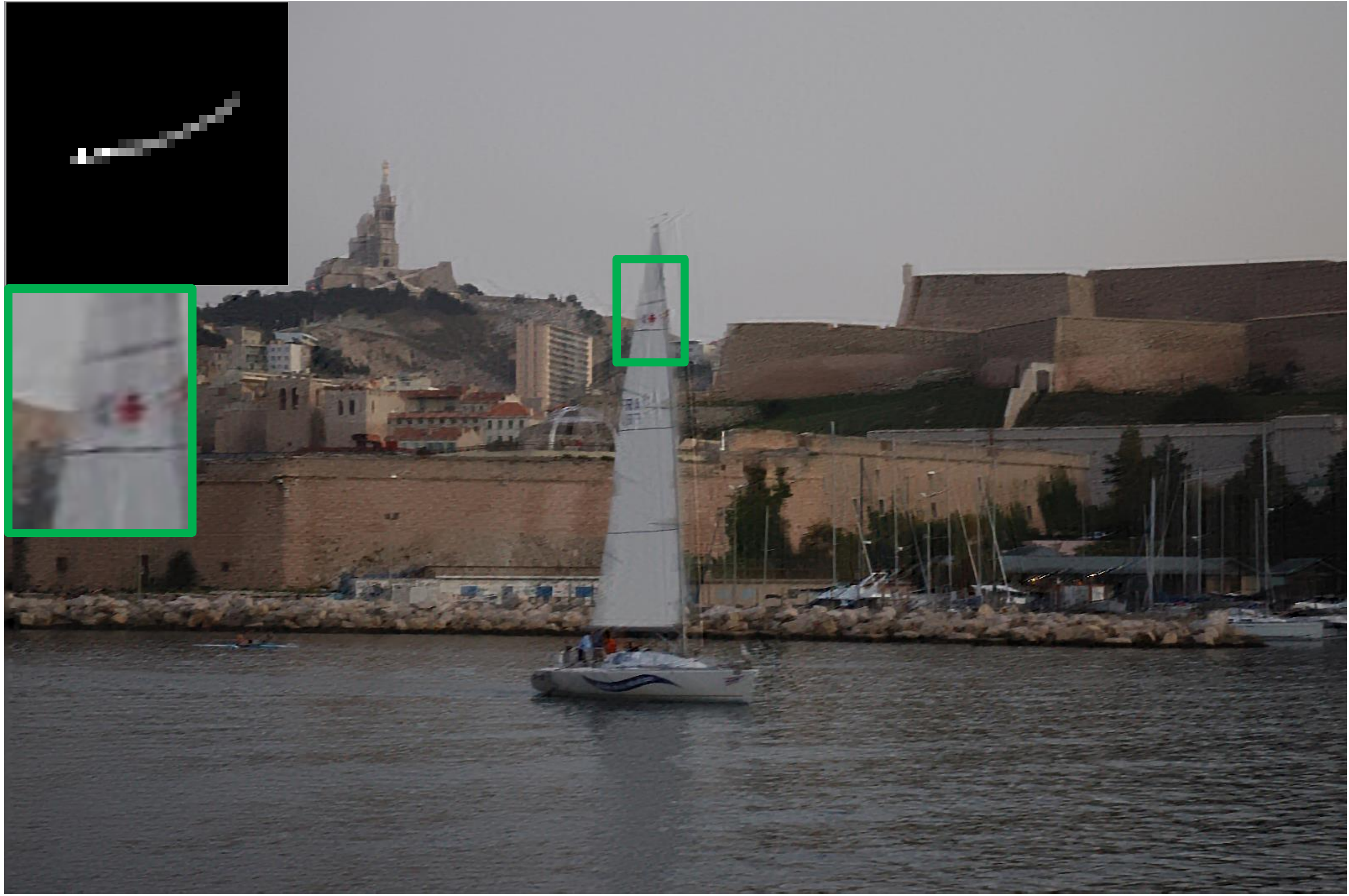}\\
\hspace{-0.3cm}
 (c) Coarse Kernel
 \hspace{-0.3cm}
& (d) Refined Kernel\\
\end{tabular}
\end{center}
\caption{\label{fig:compare}  \em (a) The blurry image from dataset \cite{pan2016blind}. (b) Deblurring results of \cite{Nah_2017_CVPR}. (c) Our deblurring result with the coarse blur kernel built from the autocorrelation of the absolute phase-only image. (d) Our deblurring result with the refined kernel. The refined kernel can better improve the deblurring result by looking at the close-up of the part of the sail with detailed sharp edges. Note that the blur kernel is zoomed in the corner.
}
\end{figure}
\subsection{Uniform Blur from Non-linear Motion} 

The blurry image is formed by the integral of light intensity over the exposure period. For more complex motion, the autocorrelation image $\calA(|P(\vB)|)$ will show more bright points representing high correlation values (see Fig. \ref{fig:fig1} (c) and Fig. \ref{fig:phase_corr} (c) for examples).

In general, in the case of uniform (spatially-invariant) blur, one may 
write $\vB = \vk \otimes \vL$, so, allowing for the possibility of noise, 
the deblurring problem (with known kernel) may be formulated as finding
$\argmin_{\vL} \left\|\vk \otimes \vL - \vB\right\|_2^2$.
In most cases, however, blurring acts as a form of low-pass filter -- high-frequency
information is lost.  Consequently, this problem is not well-conditioned. Thinking
of convolution with known $\vk$ as being a linear operator, there exist near-zero
eigenvalues whose eigenvectors correspond to high-frequency components
of the signal (image).  The deblurring process is to restore the lost frequency components
of the image.  If high-frequency components are over-emphasized in the deblurring process,
the resulting latent image $\vL$ will be noisy, or edges will show ringing. 
A common solution to this is to add
a regularization term that discourages excessive high-frequency components.
One is therefore led to the following minimization problem.
{\small
\begin{equation}\label{eq:energy1}
\min_{\vL} \left\|{\vk} \otimes \vL - \vB\right\|_2^2 + \mu_2 \,  h(\nabla\vL) ~,
\end{equation}
}
where $h(\cdot)$ is a penalty term used to discourage excessive gradients,
which are indicative of noise and over-emphasized edges.



In the case of non-linear motion, the kernel is not known exactly, but an initial value
of $\vk$ may be estimated directly from the autocorrelation of the absolute~\emph{phase-only image} as described previously.
Our final goal is to further refine the kernel $\vk$ and estimate the latent sharp image $\vL$ 
by solving
\begin{equation}\label{eq:Energy}
\min_{\vL,\vk} \left\| \vk \otimes \vL - \vB\right\|_2^2 + \mu_1 \left\| 
\vk\right\|_\mathrm{2}^2 + \mu_2 \, h(\nabla \vL) ~,
\end{equation}
where $\mu_1$ and $\mu_2$ are weight parameters.
The first term encodes the fact that the modelled blurry image should be similar to the observed image. The
second term is to regularize the solution of the blur kernel. The third term
prevents over-sharpening.

The optimization of our energy function defined in Eq.~(\ref{eq:Energy}) involves two sets of 
variables, the kernel and the latent image.  We perform the minimization 
iteratively starting with the initial estimate of $\vk$ given
by the phase-only technique.  (See Fig. \ref{fig:compare} for an example).
 
\vspace{-2 mm} 
\subsubsection{Estimating the Latent Image}
The goal is to minimize Eq.~\eqref{eq:Energy} by alternation. If $\vk$ is known, the problem comes down to minimizing 
Eq.~\eqref{eq:energy1}.

\SKIP{
Giving an estimated blur kernel $\tilde{\vk}$, the latent image $\vL$ can be solved by Eq. \eqref{eq:energy1}
{\small
\begin{equation}\label{eq:energy1}
\min_{\vL} \left\|\tilde{\vk} \otimes \vL - \vB\right\|_2^2 + \mu_2 \left\|\nabla \vL\right\|_\mathrm{0},
\end{equation}
}
Our framework contains a $L_0$ regularization term, which can effectively approximate the sparsity during the optimization. 
} 
Specifically, we use a truncated-quadratic gradient regularization term
{\small
\begin{equation}\label{eq:L0}　\nonumber
h(\nabla \vL) = \sum_{x,y} \, \min \big(
\left\| \nabla_{xy} \vL / \epsilon\right\|^2, 1 \big)
\end{equation}
}
where $\epsilon \in [0.1, 1]$ and $\nabla_{xy} \vL$ represents the gradient of $\vL$
at image coordinates $(x, y)$.  This regularization term smooths out small noise,
while allowing occasional large gradients (intensity differences).
This type of term, proposed by \cite{Blake-Zisserman} 
is widely used to regularize noise and gradients 
in stereo \cite{Veksler} and was also used in deblurring
in \cite{xu2013unnatural}). 
Because the truncated quadratic is non-convex, the optimization problem is non-convex.
We use the method of half quadratic splitting, as in ~\cite{xu2011image}, 
to minimize this cost function,
though other methods such as Iterative Reweighted Least Squares could be used
for such truncated-quadratic cost \cite{Aftab-Hartley:WACV}.

\SKIP{
To solve the non-convex energy minimization problem \eqref{eq:Energy}, we use the half
quadratic splitting $l_0$ minimization approach~\cite{xu2011image}. The energy
function can be written as 
\begin{equation}\label{eq:EstL}
\min_{{\vL}} \left\| \tilde{\vk} \otimes {\vL} - \vB\right\|_2^2
+ \mu_2 \left( \frac{1}{\epsilon^2}  \left\| \nabla {\vL} -
\phi(\vL) \right\|^2 + \left\| \phi({\vL})
\right\|_0 \right),
\end{equation}
where $\phi({\vL}) $ corresponds to image gradients in the
horizontal and vertical directions, and $\left\| \phi({\vL})
\right\|_0$ is the zero-norm.
\begin{equation}\nonumber
\phi(\vL) = 
\begin{cases}
0 & \mbox{ if } | \nabla {\vL} | \leq \epsilon\\
\nabla {\vL} & \mbox{otherwise}
\end{cases}.
\end{equation}

} 
\vspace{-2 mm}
\subsubsection{Refining the Kernel}


Now, with ${\vL}$ known, the motion blur kernel can be refined by solving 
{\small
\[
\min_{\vk} \left\|  \vk \otimes {\vL} - 
\vB\right\|_2^2 + \mu_1 \left\| \vk\right\|_\mathrm{2}^2 ~.
\]
 }
This is a quadratic problem, and can be solved directly by taking gradients,
which results in a set of linear equations.
More efficiently, we solve it in the Fourier domain, in which case there is
a closed-form solution
{\small
\[
\calF(\vk) = \overline{\calF(\vL)} \odot \calF(\vB)~\big/~
\big(\overline{\calF(\vL)} \odot \calF(\vL) + \mu_1\big) ~,
\]
}
where the division is carried out point-wise (as are the multiplications).
Then $\vk$ is found by the inverse transform, and then normalized to sum to $1$.

The algorithm alternates between recomputing $\vL$ and $\vk$ until convergence, or for
a fixed number of steps.


\vspace{-2 mm}
\section{Extension to Non-uniform Deblurring}

\begin{figure}[t]
\begin{center}
\begin{tabular}{cccc}
\hspace{-0.20 cm}
\includegraphics[width=0.225\textwidth]{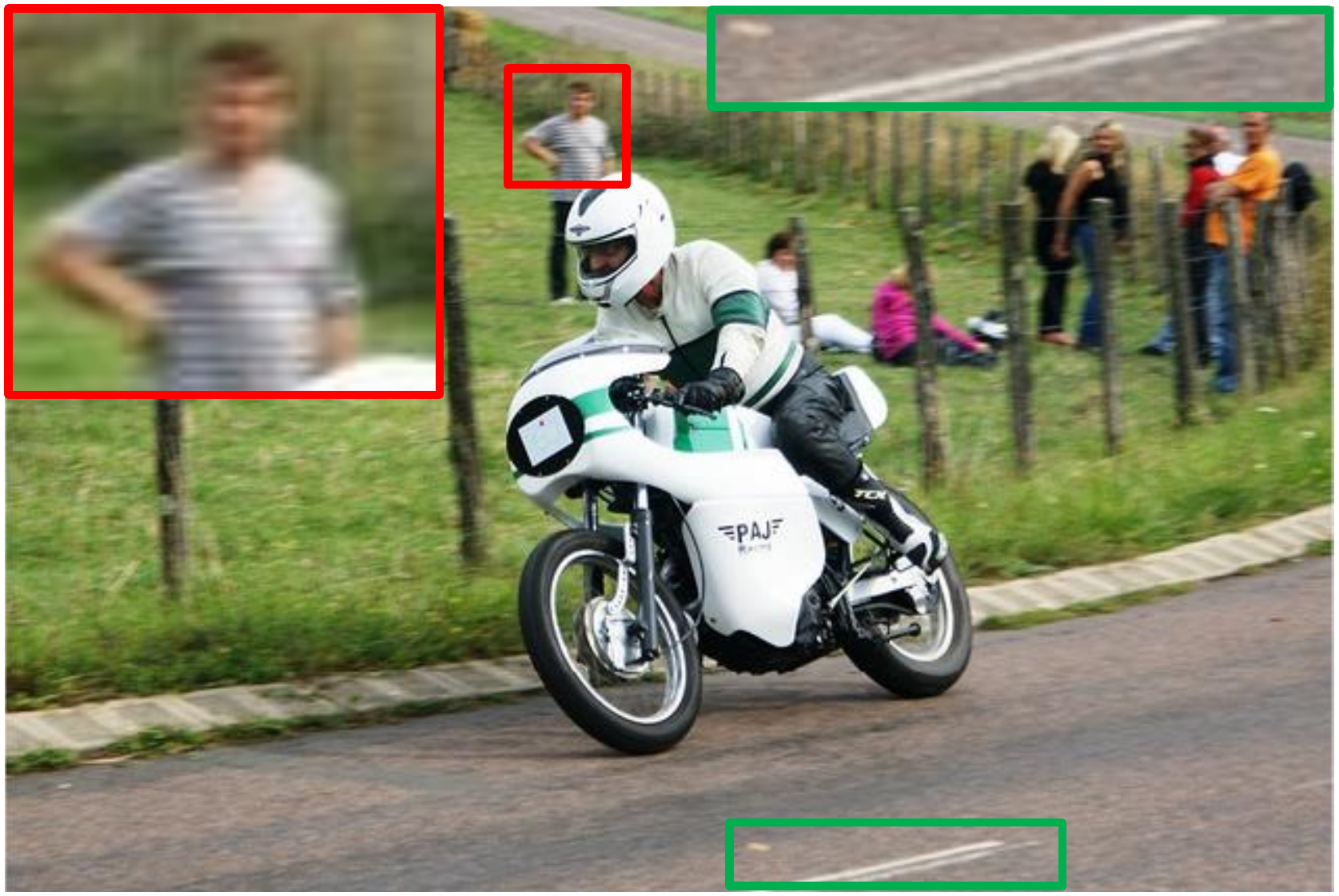}
\hspace{-0.30 cm}
&\includegraphics[width=0.225\textwidth]{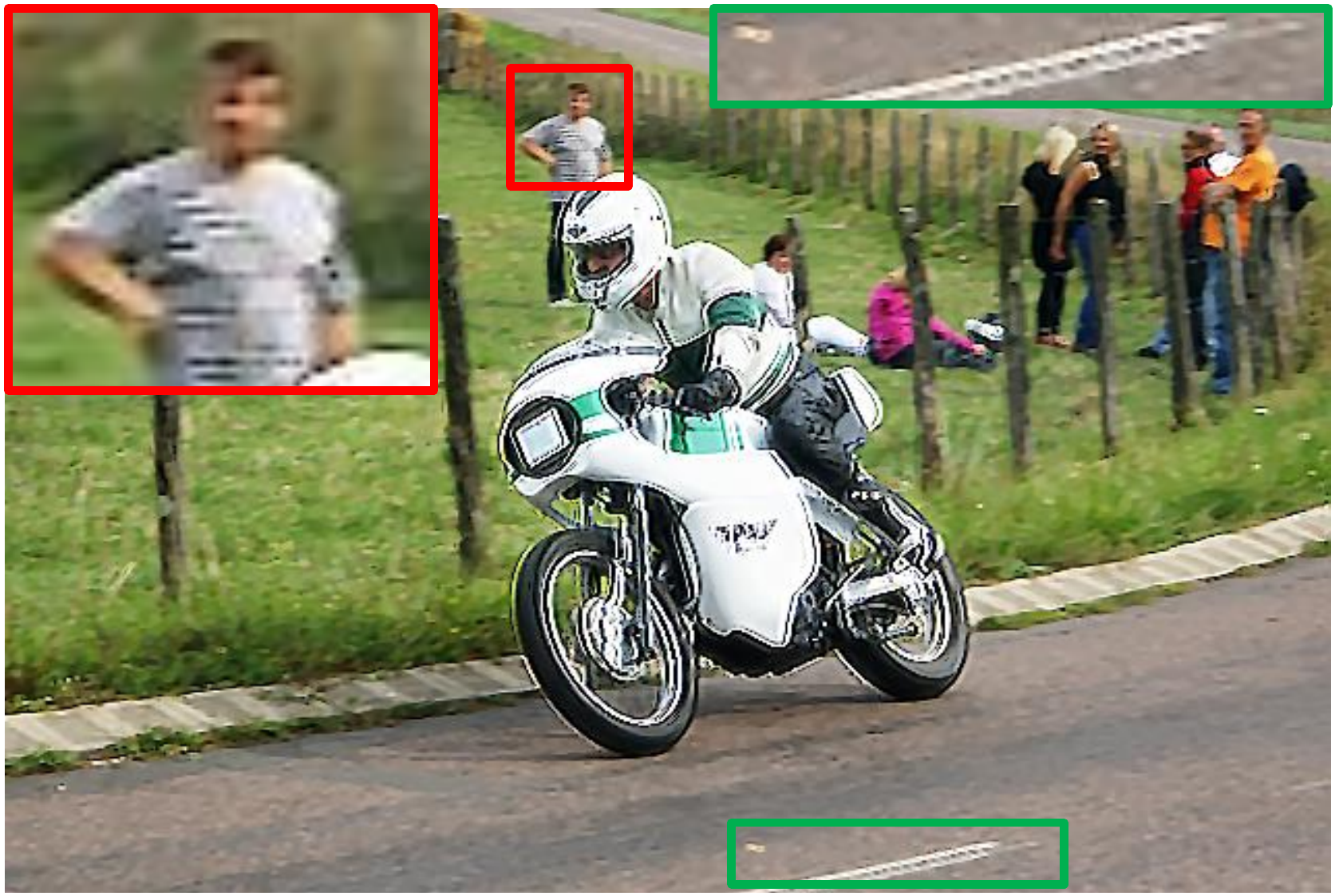}\\
\hspace{-0.20 cm}
(a) Blurry Image  
\hspace{-0.30 cm}
& (b) Ours (Uniform) \\
\hspace{-0.20 cm}
\includegraphics[width=0.225\textwidth]{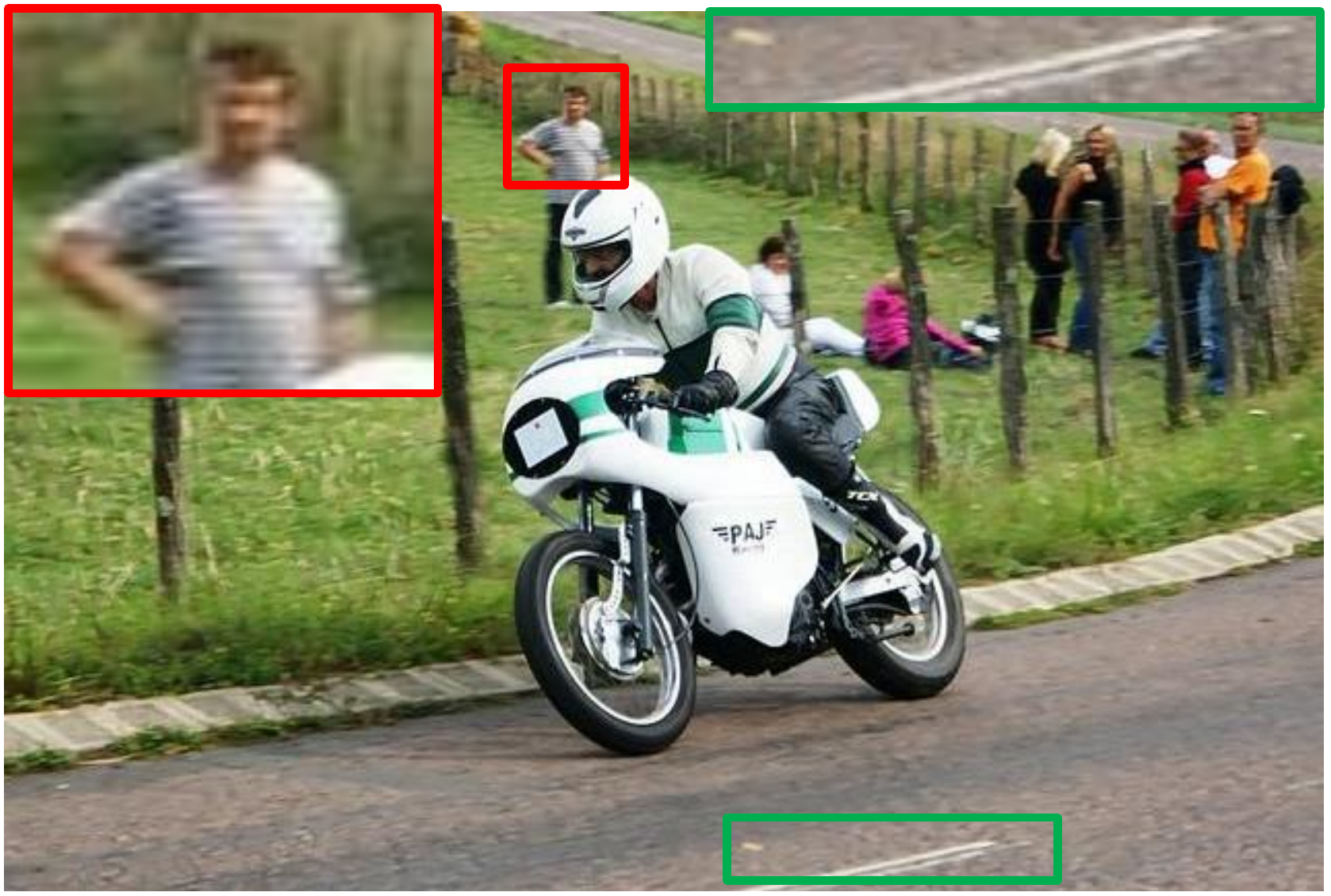}
\hspace{-0.30 cm}
&\includegraphics[width=0.225\textwidth]{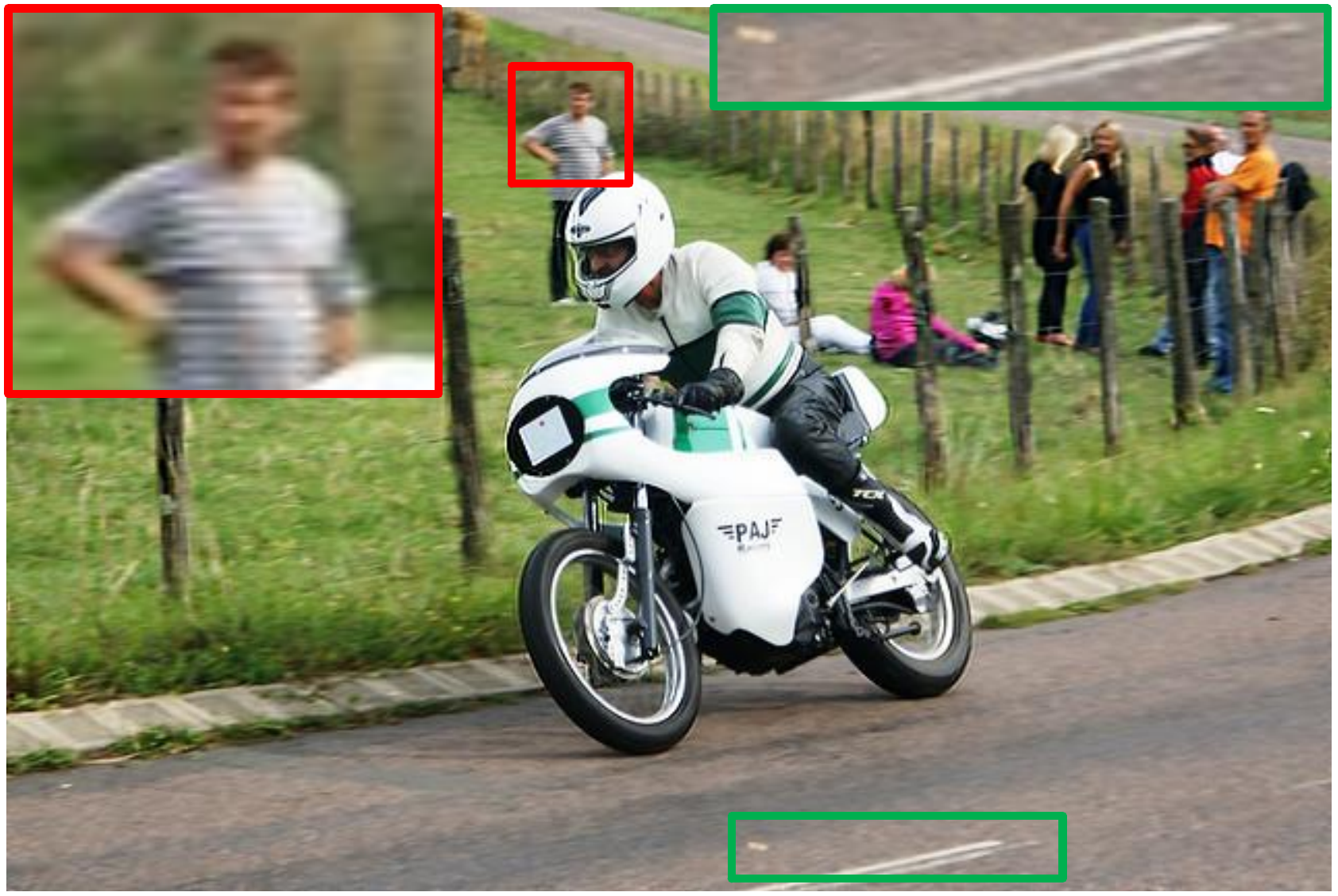}\\
\hspace{-0.20 cm}
(c) Nah \cite{Nah_2017_CVPR}
\hspace{-0.30 cm}
& (d) Gong \cite{gong2017motion} \\
\hspace{-0.20 cm}
\includegraphics[width=0.225\textwidth]{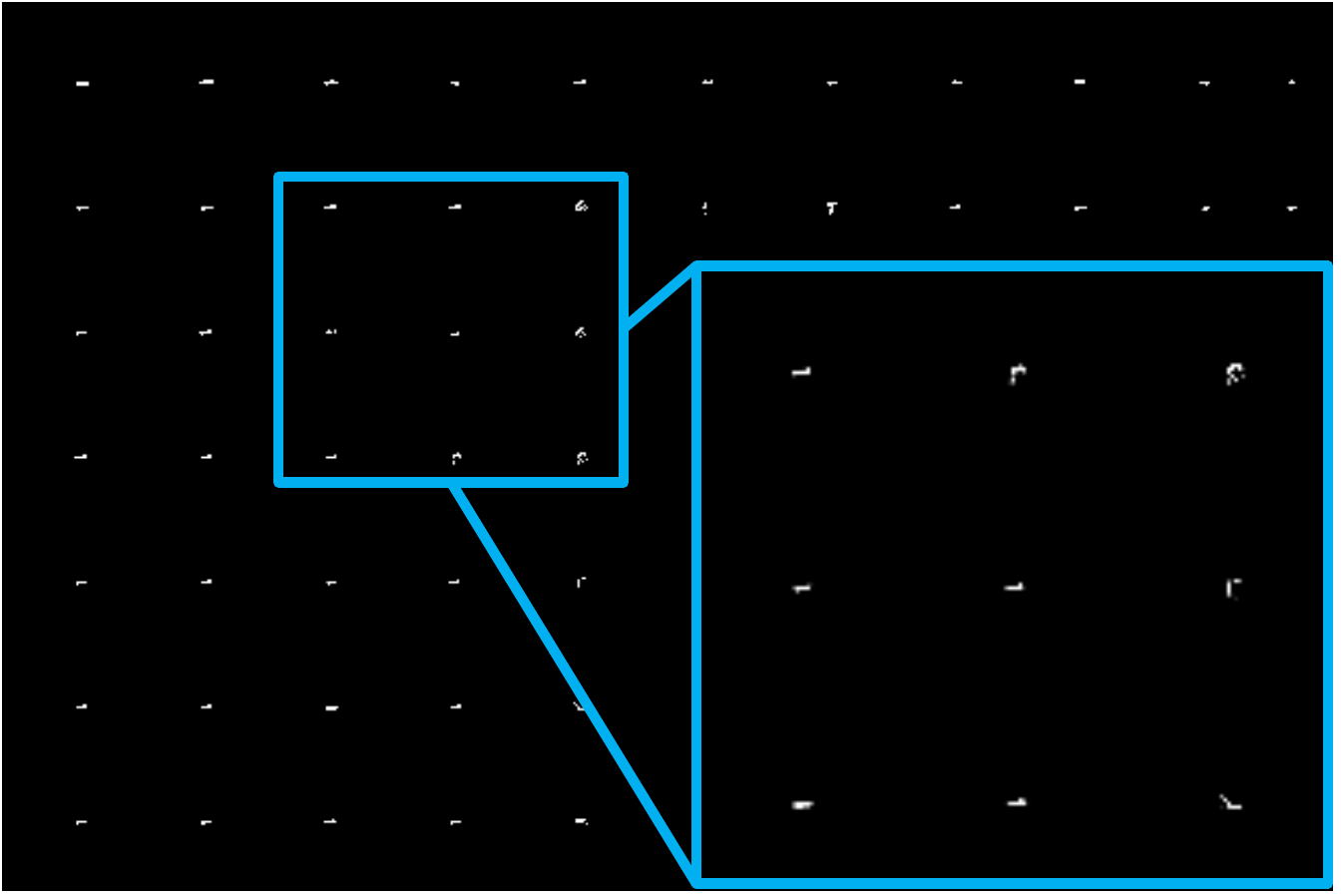}
\hspace{-0.30 cm}
&\includegraphics[width=0.225\textwidth]{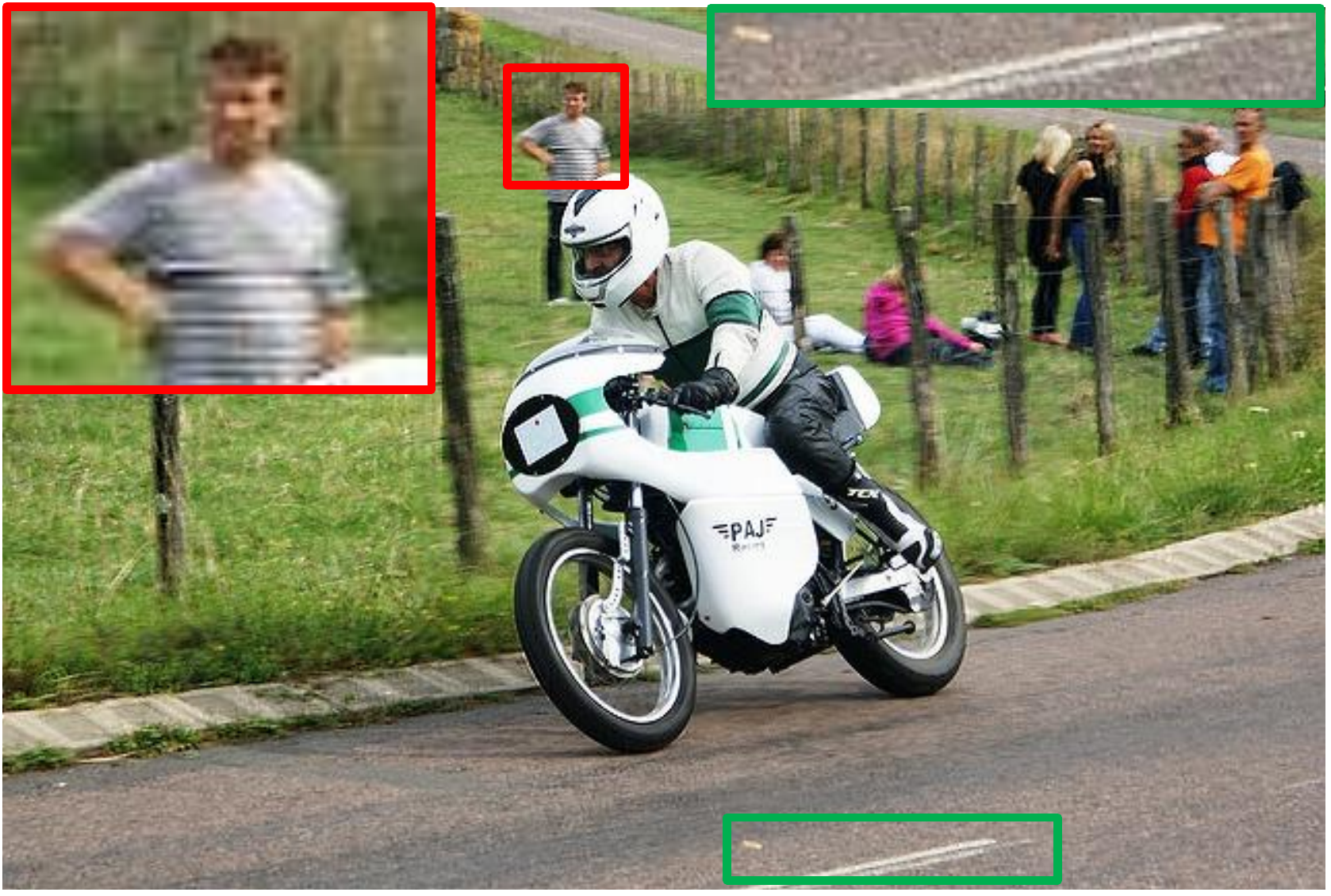}\\
\hspace{-0.20 cm}
(e) Blur Kernel
\hspace{-0.30 cm}
& (f) Ours (Non-uniform)\\
\end{tabular}
\end{center}
\vspace{-1.5 mm}
\caption{\em \label{fig:nonkernel}  Example of our non-uniform blur kernel where the real blurry image is from \cite{gong2017motion}. 
(a) Input blurry image. 
(b) Our deblurring
results by using uniform blur model and its blur kernel. We can see clearly that
the man in a plaid shirt seems not deblurred because of the improper kernel. 
(c) Deblurring result of \cite{Nah_2017_CVPR}.
(d) Deblurring result of \cite{gong2017motion}.
(e) Non-uniform blur kernel. 
(f) Our deblurring result by using non-uniform blur
model and kernel. 
}
\end{figure}

Our method can be easily extended to handle non-uniform blur (\eg, the background and
foreground undergo different blur) by deblurring the image patch-by-patch or
layer-by-layer. Each patch or layer of the image corresponds to a different blur kernel.
The new non-uniform blur model can be expressed as
{\small
\begin{equation}\label{eq:nonblurmodel}
\vB = \sum_{i=1}^N {\vk}_i \otimes {\vl}_i,
\end{equation}
}
where $N$ denotes the number of segmented patches or layers, $\vl_i={\bf M}_i\odot\vL$ is to extract the $i$-th patch or layer of the latent image, ${\bf M}_i$ is a binary mask with non-zeros values in the region corresponding to the $i$-th patch or layer in $\vL$, and $\vk_i$ denotes the blur kernel corresponding to the $i$-th patch. Similary, we define $\vB_i=\vk_i\otimes \vl_i$ and $\vB=\sum_{i=1}^N\vB_i$.
Each layer can be handled using our proposed uniform deblurring approach in Section~\ref{sec:uniformDeblurring}. The final latent image $\vL$ is $\sum_{i=1}^{N}{\vl_i}$.  In Fig.~\ref{fig:nonkernel}, we give an example of the deblurring results for uniform and non-uniform blur models. The image is a real blurry image from dataset \cite{gong2017motion}. Clearly, our non-uniform deblurring achieves better results than our uniform-deblurring model and the other existing non-uniform deblurring methods which either use additional depth, camera pose information~\cite{hu2014joint,gupta2010single,whyte2012non} or use deep convolutional neural networks~\cite{gong2017motion, Nah_2017_CVPR}.
\section{Experiment}\label{sec:experiments}

\begin{figure}[!t]
\begin{center}
\includegraphics[width=0.395\textwidth]{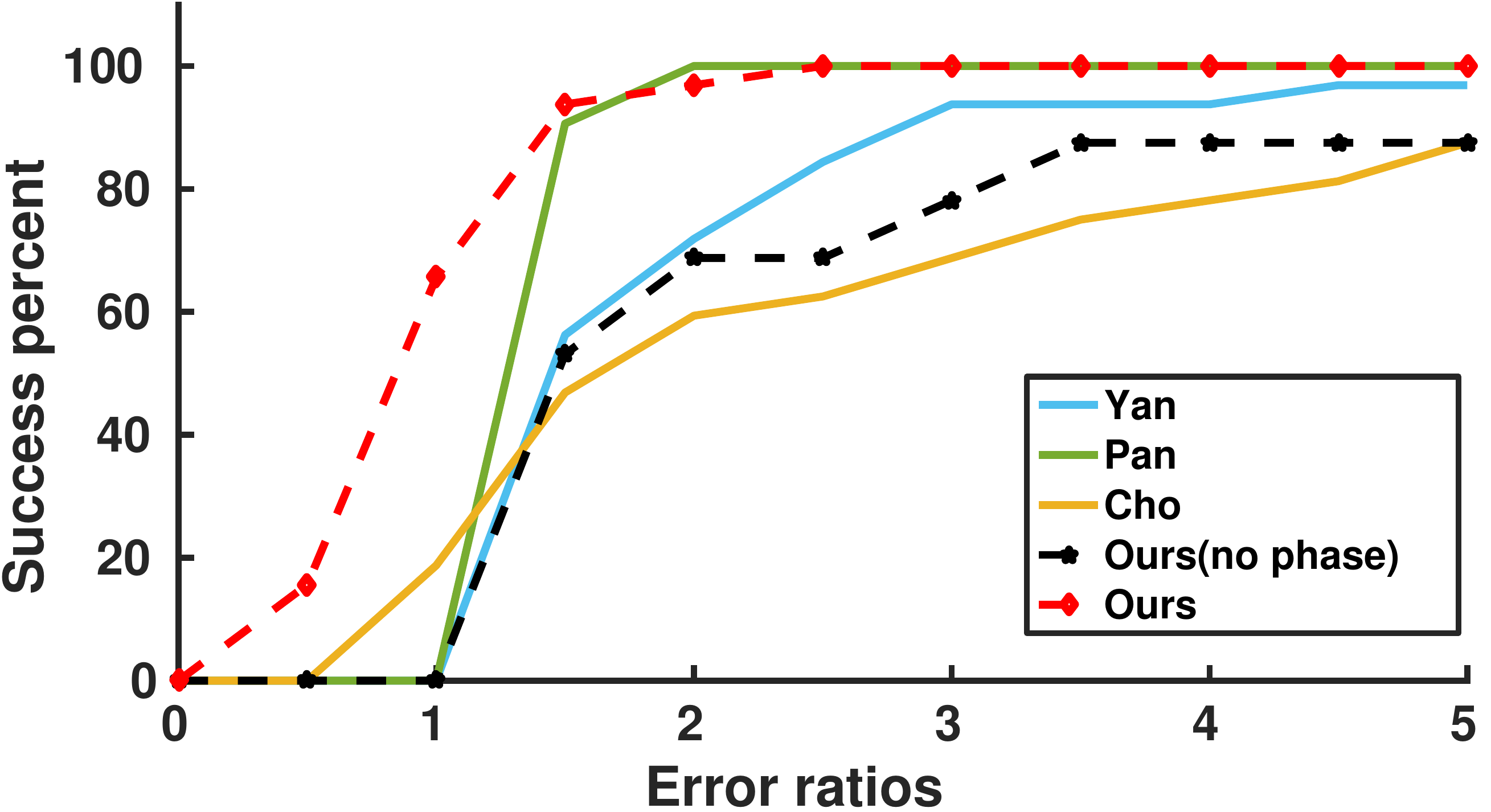}
\end{center}
\vspace{-2 mm}
\caption{\em \label{fig:levin}Quantitative evaluations on dataset \cite{levin2009understanding}.
 We report the experimental results with and without using the blur kernel estimated from the {\it phase-only image} (\emph{'Ours(no phase)'}). The results further demonstrate the effectiveness of blur kernel estimation from the {\it phase-only image}.
 }
\end{figure}
{\small
\begin{table}\footnotesize
\centering
\caption{\em Quantitave comparison on the dataset~\cite{levin2009understanding}.}
\label{errorratio}
\begin{tabular}{c|c|c|c|c|c}
\hline
          & Cho \cite{cho2011handling}  & Pan \cite{pan2016blind}  & Yan \cite{yan2017image}   & \begin{tabular}[c]{@{}c@{}}Our\\ (no phase)\end{tabular}   & Our\\ \hline
 PSNR(dB) & 25.63           & 27.54          & 24.70     & 25.74             & {\bf 28.38}\\ \hline
 SSIM     & 0.7907          & 0.8626         & 0.8760    & 0.7842            &{\bf 0.9250}\\ \hline
 SSD      & 2.6688          & 1.2747         & 1.6802      & 3.2517          & {\bf 0.8776}  \\ \hline
\end{tabular}
\end{table}
}

\subsection{Experimental Setup}
\begin{figure*}[!t]
\vspace{-0.2cm}
\begin{center}
\begin{tabular}{cccc}
\hspace{-0.3cm}
\includegraphics[width=0.200\textwidth]{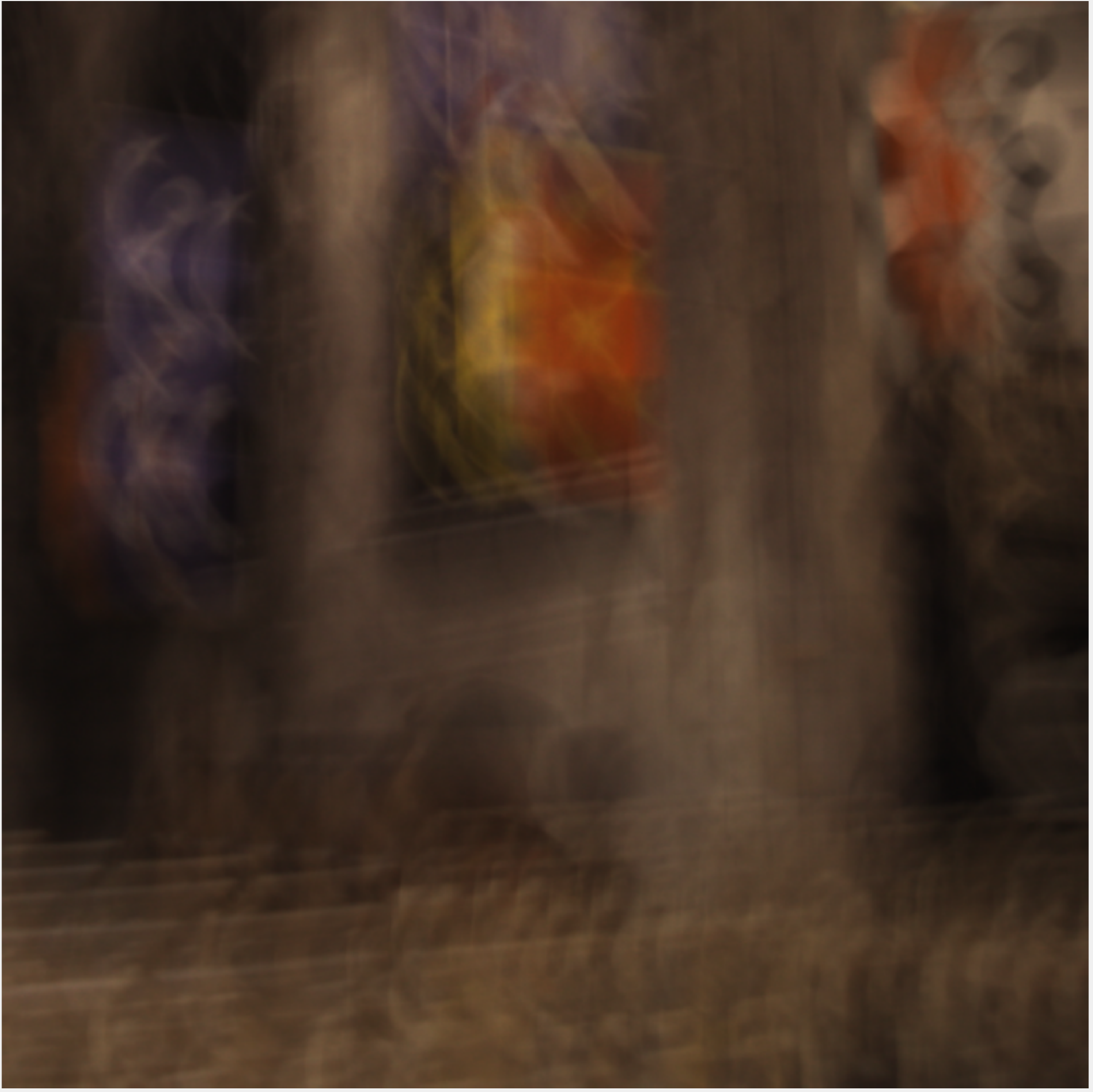}
&\includegraphics[width=0.200\textwidth]{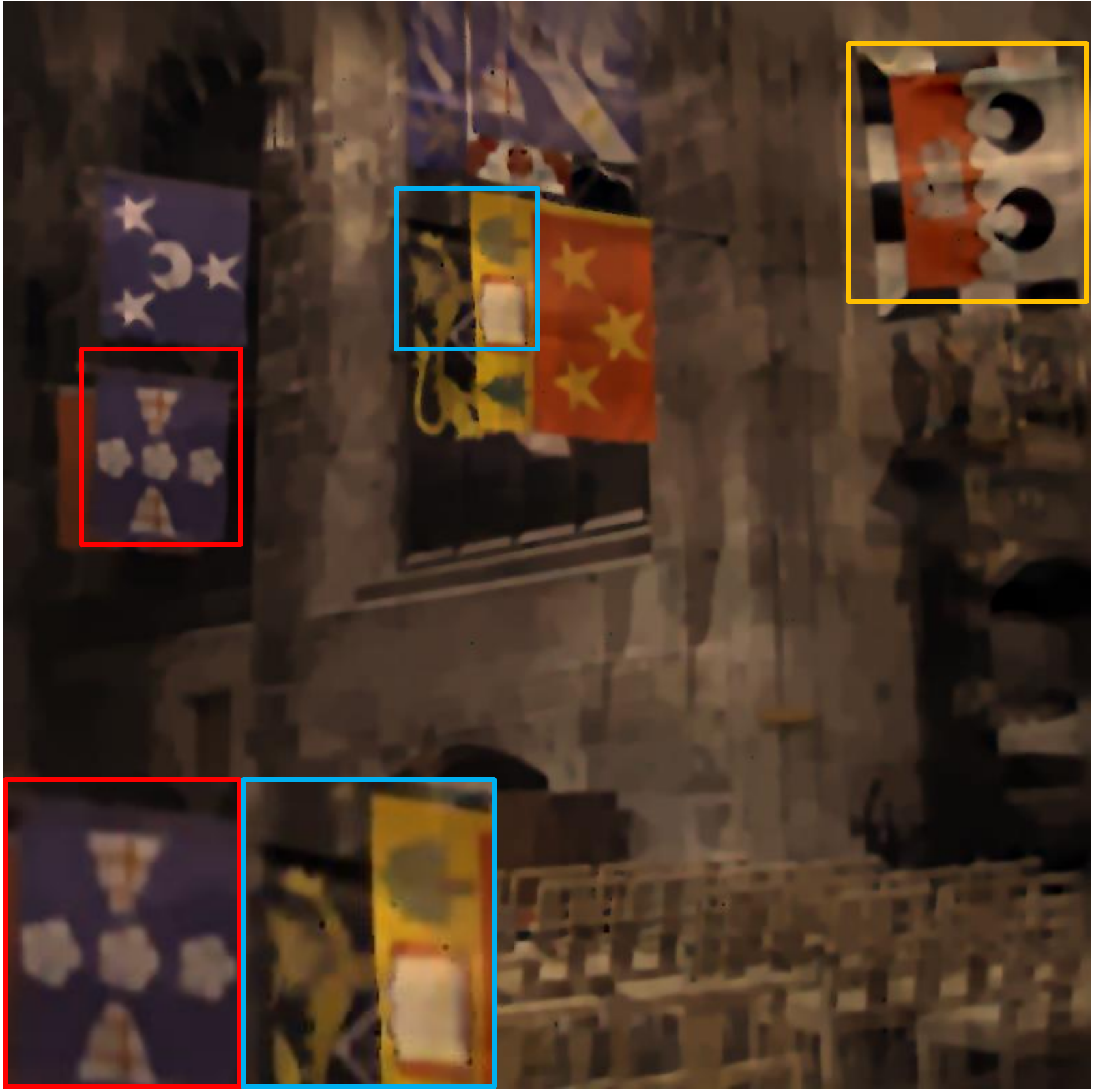}
&\includegraphics[width=0.200\textwidth]{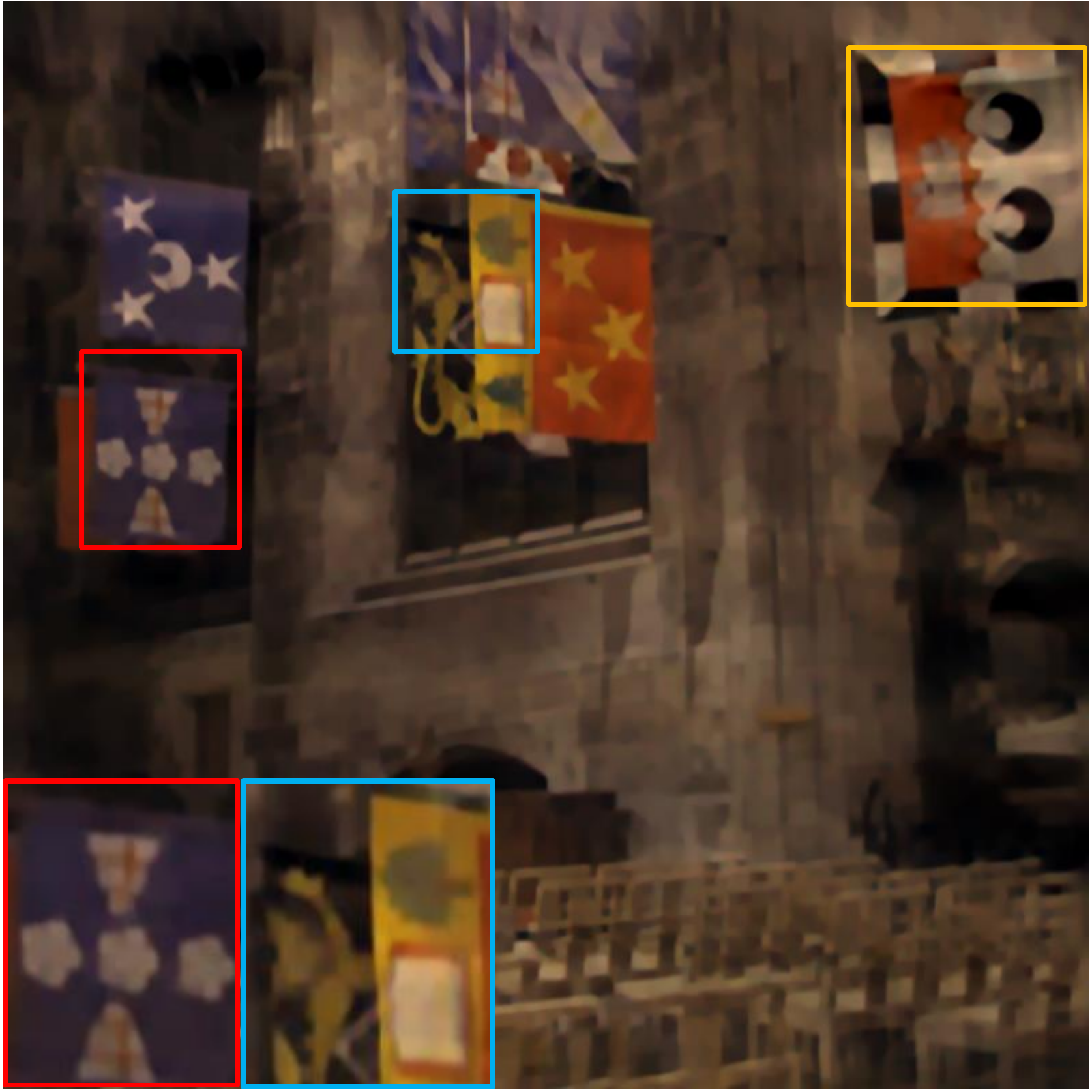}
&\includegraphics[width=0.200\textwidth]{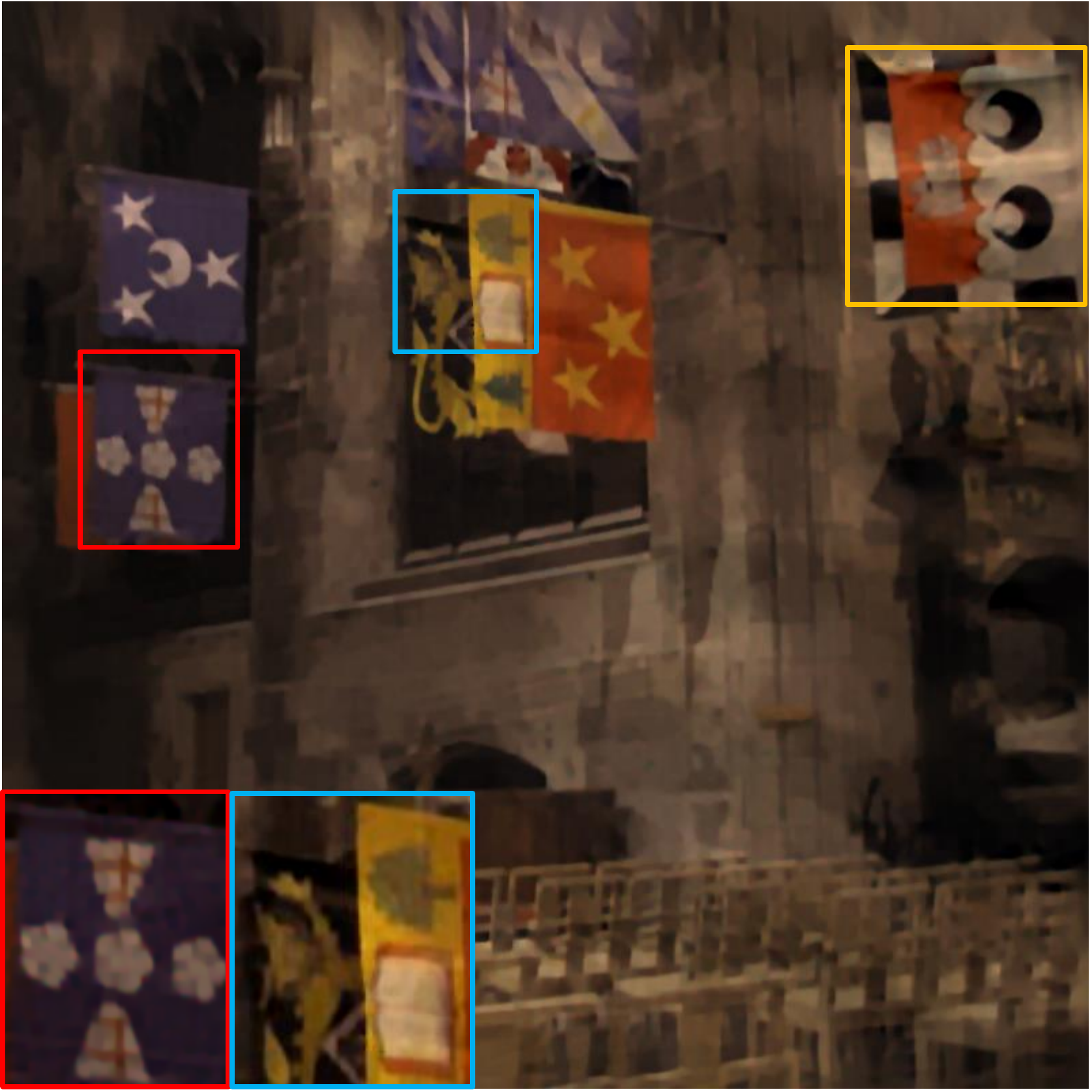}\\
\hspace{-0.4cm}
\includegraphics[width=0.200\textwidth]{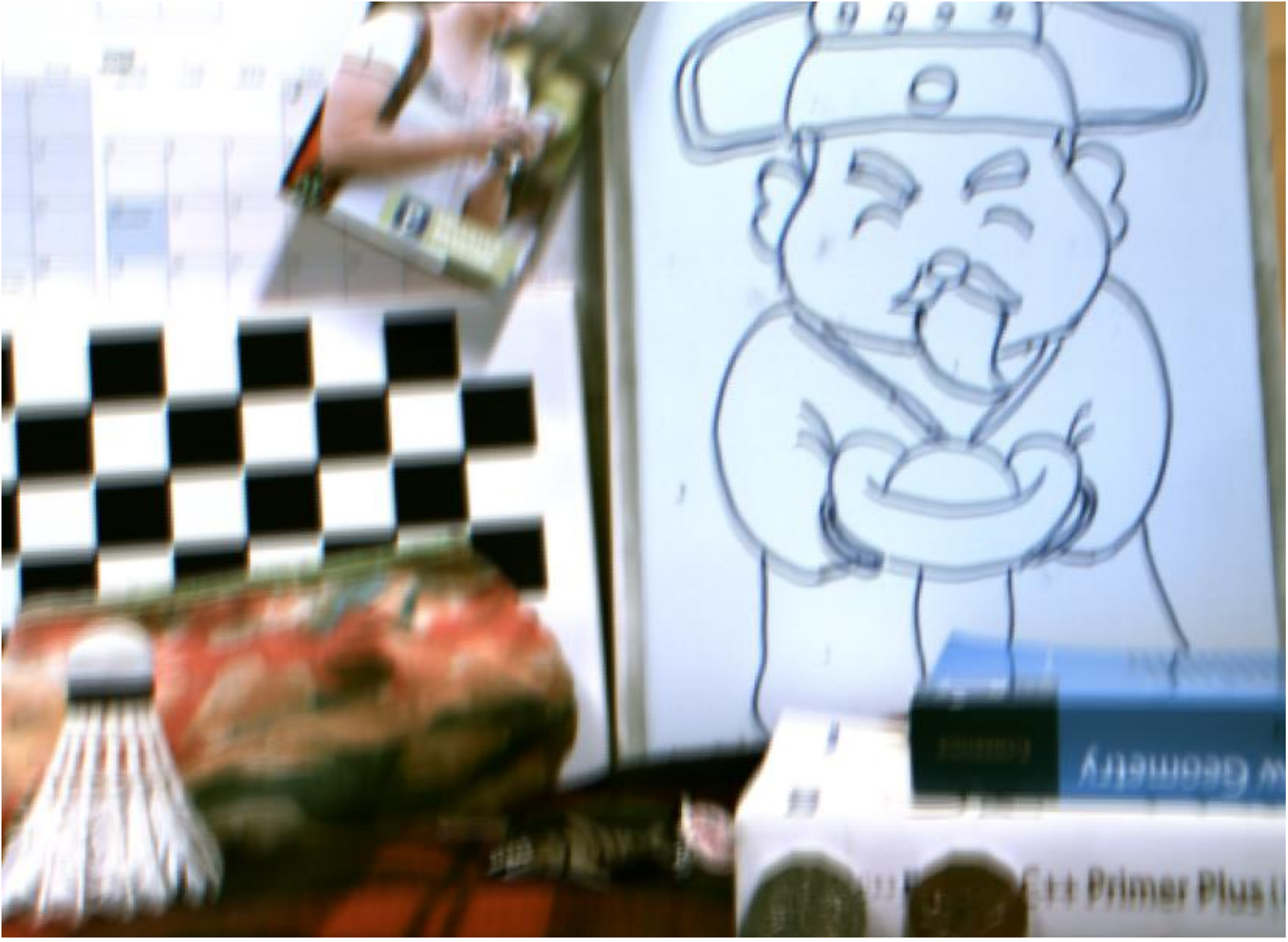}
&\includegraphics[width=0.200\textwidth]{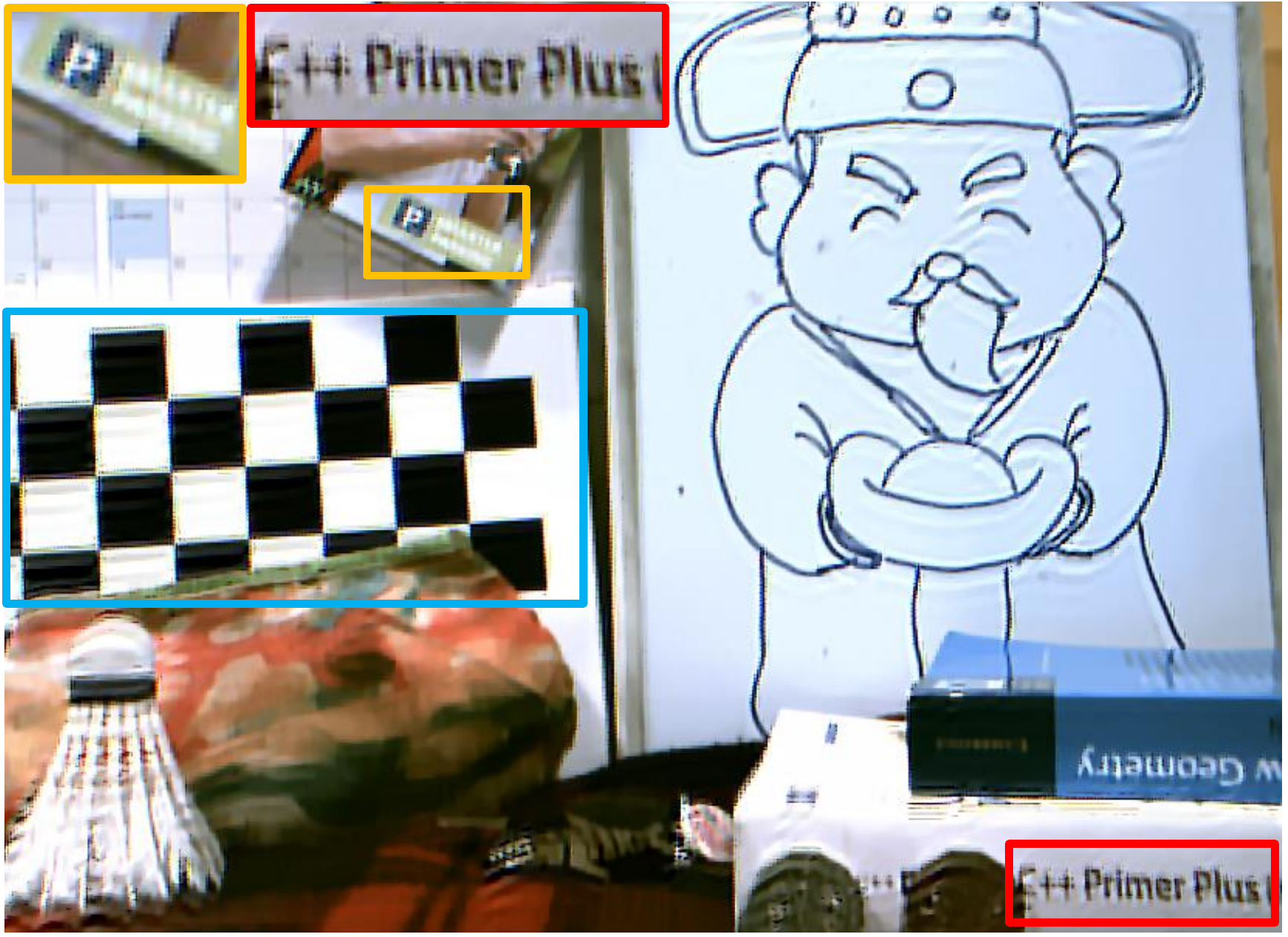}
& \includegraphics[width=0.200\textwidth]{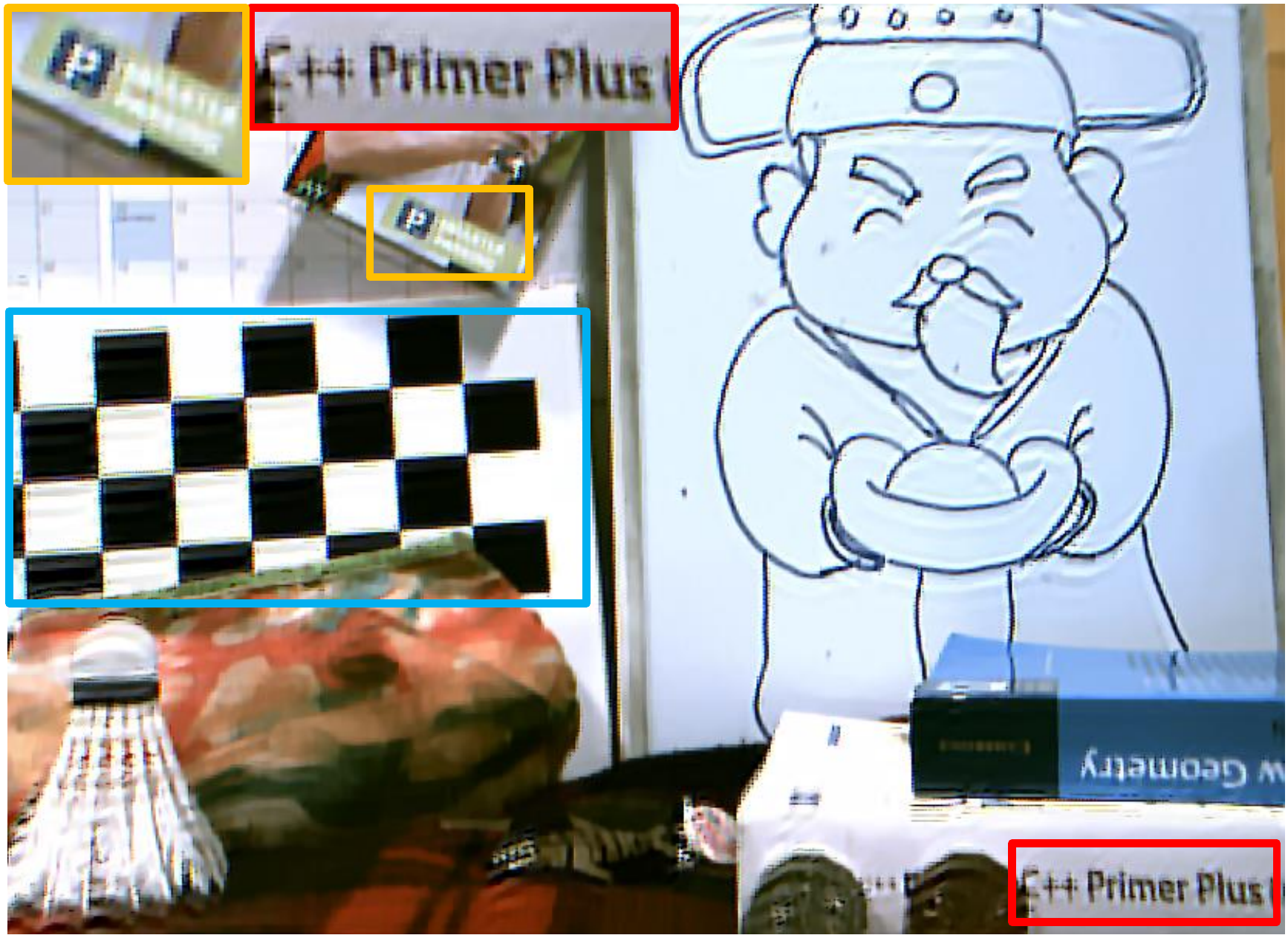}
&\includegraphics[width=0.200\textwidth]{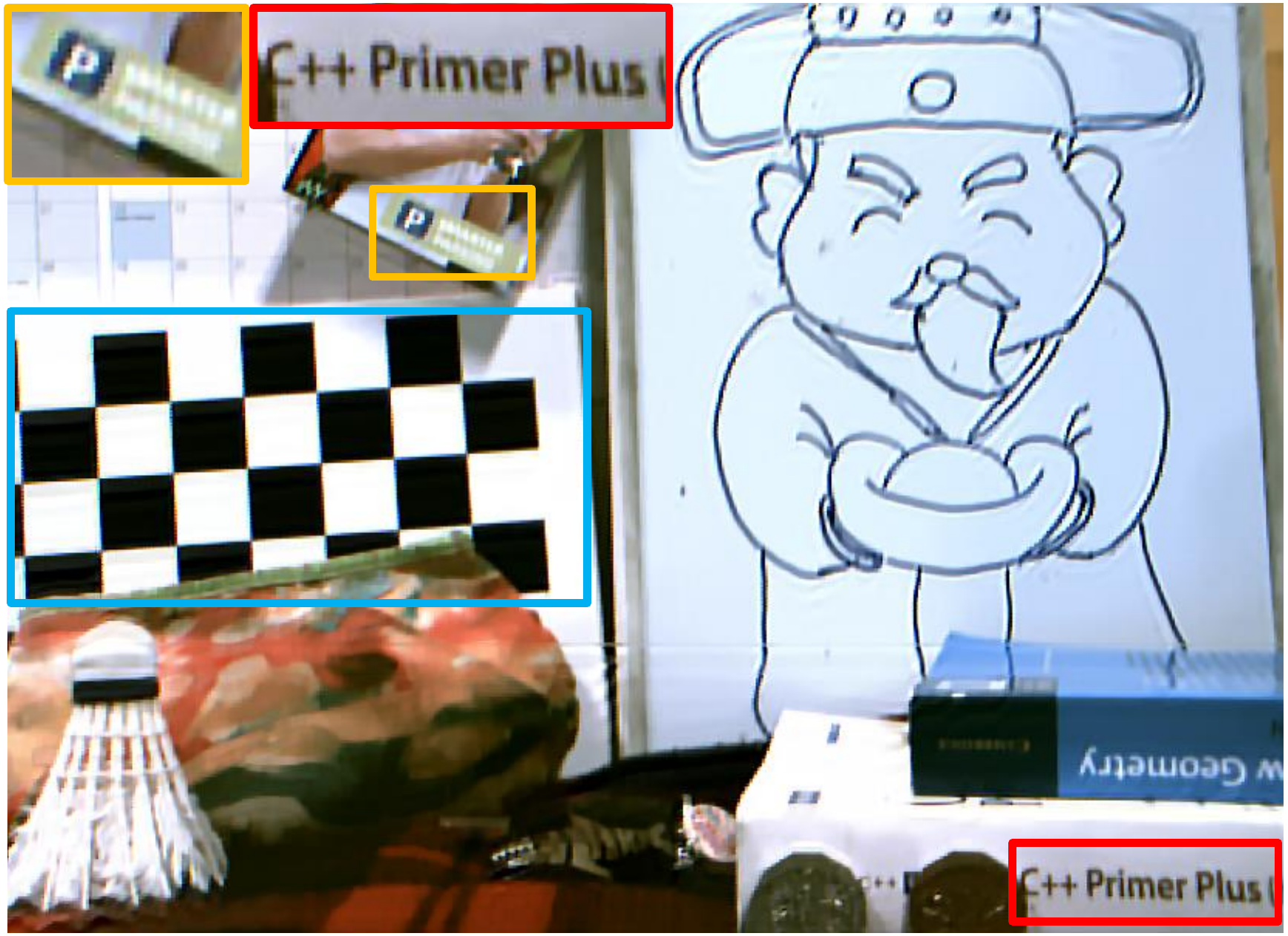}\\
\hspace{-0.3cm}
\includegraphics[width=0.200\textwidth]{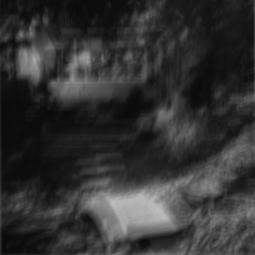}
&\includegraphics[width=0.200\textwidth]{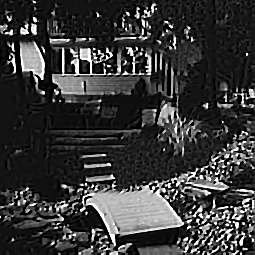}
&\includegraphics[width=0.200\textwidth]{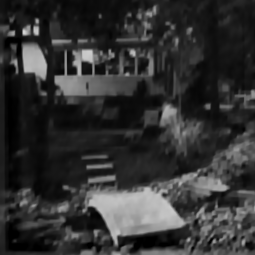}
&\includegraphics[width=0.200\textwidth]{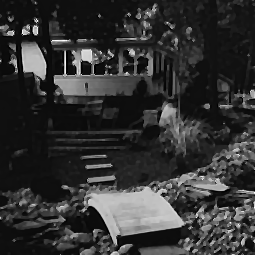}\\
\hspace{-0.3cm}
(a) Blurry Image  
&(b) Yan \cite{yan2017image} 
&(c) Pan \cite{pan2016blind}  
&(d) Ours\\
\end{tabular}
\end{center}
\vspace{-2 mm}
\caption{\label{fig:1_12}\em Qualitative comparison on example images from dataset~\cite{kohler2012recording}(top),~\cite{levin2009understanding}(bottom) and image taken by ourselves (middle).~(a) Input blurry images. (b) Deblurring results of~\cite{yan2017image}. (c) Deblurring results of~\cite{pan2016blind}. (d) Our deblurring result. (Best viewed on screen).
}
\end{figure*}

\vspace{-2mm}
\begin{table*}\footnotesize
\centering
\caption{\em Quantitative comparisons on the 
dataset~\cite{kohler2012recording}, where \cite{Nah_2017_CVPR,DeblurGAN} are deep based methods.}
\label{all_all}
\hspace*{-0.2cm}
\begin{tabular}{c|c|c|c|c|c|c|c|c}
\hline
           & Blurry Image  & Whyte \etal \cite{whyte2012non}& Xu \etal \cite{xu2013unnatural}  & Pan \etal \cite{pan2016blind}  & Yan \etal \cite{yan2017image}  & Nah \etal \cite{Nah_2017_CVPR}  & Kupyn \etal \cite{DeblurGAN} & Ours    \\ \hline
 PSNR(dB)  &  24.93    & 27.03  & 27.47    & 29.95    & 28.42  & 26.48  & 26.10  & \bf{30.18}       \\ \hline
 SSIM      & 0.783     & 0.809  & 0.811    & 0.932    & 0.897  & 0.807  & 0.816 &\bf{0.933}      \\ \hline
\end{tabular}
\end{table*}
\noindent{\bf{Dataset.}} We evaluate our approach on the datasets provided by~\cite{kohler2012recording,pan2016blind,sturm2012benchmark,gong2017motion,levin2009understanding}
and images captured by ourselves, which covers images from man-made scene,
natural scene and images containing text (see Fig.~\ref{fig:compare}, \ref{fig:nonkernel}, \ref{fig:1_12} for examples).

\noindent{\bf{Baselines and evaluation metric.}} 
Since our proposed approach can handle both uniform and non-uniform blurs, we compare with state-of-the-art methods for both cases separately. For traditional methods (non-deep learning methods), we compare with
~\cite{yan2017image,pan2016blind,cho2009fast,whyte2012non,xu2013unnatural}. For deep learning based methods, we compare with \cite{gong2017motion,Nah_2017_CVPR,DeblurGAN} which can handle spatially-variant blur. We report the PSNR, SSIM on datasets~\cite{levin2009understanding,kohler2012recording}
and~\emph{error ratio}\footnote{~\emph{Error ratio} is introduced in~\cite{levin2009understanding} which measures the ratio between the SSD (Sum of Squared Distance) of the deconvolution error computed with the estimated kernel and the ground truth kernel.} on dataset~\cite{levin2009understanding} which provides the ground truth blur kernels for evaluation. 

\noindent{\bf{Implementation details.}}  We validate the parameters in our model 
on three reserved images for each dataset and use coarse-to-fine strategy for deblurring.
We set $\mu_1 = 2$, $\mu_2 = 0.005$ for our experiment. 
Our framework is implemented using MATLAB with C++
wrappers. It takes around 40 second to process one image ($ 800 \times 800$) on a single i7 core running at 3.6 GHz. 
\vspace{-0.1cm}
\subsection{Experimental Results}
\vspace{-0.1cm}
The dataset introduced in~\cite{levin2009understanding} is a widely used uniform blur dataset, which contains 32 blurry images generated by 4 ground truth images and 8 blur kernels. We perform the quantitative and qualitative evaluation on this dataset. Results are shown in Fig.~\ref{fig:levin}, \ref{fig:1_12} and Table \ref{errorratio}, which demonstrates that our proposed approach achieves competitive results.

The~\emph{Natural dataset} is generated by \cite{kohler2012recording} with
camera motion measured and controlled by a Vicon tracking system.
Specifically, the dataset provides blurry image, its latent image,
and ground truth blur kernel, which allows the quantitative comparison of our approach with baselines.
The captured images are of size $800 \times 800$. 
In Table~\ref{all_all}, we show the quantitative comparison with the state-of-the-art Single-image deblurring approaches on dataset~\cite{kohler2012recording}. 
It demonstrates that our approach can achieve the best performance on the PSNR and SSIM score.

We further show the corresponding qualitative comparison results on example images in~\cite{kohler2012recording} in Fig. \ref{fig:1_12}. It clearly shows that our approach can recover more sharp details and with less ringing artifacts than other approaches, which are highlighted in the presented results. 
We also report our deblurring result in Fig.~\ref{fig:fig1}, \ref{fig:phase_corr}, \ref{fig:compare} and \ref{fig:nonkernel}, respectively. Note that our deblurring results can recover the color more faithfully than the
baselines.

\vspace{-2mm}
\section{Conclusions}
\vspace{-2mm}
Our proposed~\emph{phase-only image} based kernel estimation approach is simple (implemented in a few lines of code). The resulted image deblurring algorithm achieves better quantitative results (using PSNR, SSIM, and SSD), than the state-of-the-art methods by extensive evaluation on the benchmark datasets.~While our approach can handle the general blur cases, it still suffers from low lighting condition like other deblurring methods. Our future work will explore how to remove blurs less sensitive to lighting conditions.

\vspace{-0.1cm}
\section*{Acknowledgement}
\vspace{-0.1cm}
This research was supported in part by Australia Centre for Robotic Vision (CE140100016), the Australian Research Council grants (DE140100180, DE180100628) and the Natural Science Foundation of China grants (61871325, 61420106007, 61671387, 61603303). 
{\small
\bibliographystyle{ieee}
\bibliography{Phase-only-cvpr.bib}
}

\end{document}